%% file: main.tex
\documentclass[10pt,twocolumn,letterpaper]{article}

\usepackage{iccv}              % To produce the CAMERA-READY version

\usepackage{times}
\usepackage{graphicx}
\usepackage{amsmath}
\usepackage{amssymb}
\usepackage{booktabs}
\usepackage{adjustbox}
\usepackage{pifont}
\usepackage{multirow}
\usepackage{arydshln}
\usepackage{amsfonts}
\usepackage[dvipsnames]{xcolor}
\usepackage[outercaption]{sidecap}    
\usepackage{color, colortbl}
\usepackage[title]{appendix}
\usepackage{enumitem}
\definecolor{LightGray}{rgb}{0.92,0.92,0.92}
\definecolor{Red}{rgb}{1.0, 0.13, 0.32}
\usepackage{listings}
\usepackage[ruled]{algorithm2e}
\newcommand{\mycc}{\cellcolor{LightGray}}

\usepackage{subcaption}

\input{math_commands}

\usepackage[pagebackref=true,breaklinks=true,colorlinks,bookmarks=false]{hyperref}

\usepackage[capitalize]{cleveref}
\crefname{section}{Sec.}{Secs.}
\Crefname{section}{Section}{Sections}
\Crefname{table}{Table}{Tables}
\crefname{table}{Tab.}{Tabs.}

\makeatletter
\newcommand\notsotiny{\@setfontsize\notsotiny{6.31415}{7.1828}}
\makeatother

\iccvfinalcopy % *** Uncomment this line for the final submission

\def\iccvPaperID{6957} % *** Enter the ICCV Paper ID here

\ificcvfinal\fi

\begin{document}

%%%%%%%%% TITLE
\title{
CLIPTER: Looking at the Bigger Picture in Scene Text Recognition
\vspace{-0.3cm}
}

\author{Aviad Aberdam\textsuperscript{1}\thanks{Corresponding author {\tt aaberdam@amazon.com}.}%\\
% AWS AI Labs\\
% {\tt\small aaberdam@amazon.com}
~\quad
David Bensa\"id\textsuperscript{2}\thanks{Work done during an Amazon internship.}%\\
% Technion, Israel\\
% {\tt\small dben-said@campus.technion.ac.il}
~ \quad
Alona Golts\textsuperscript{1}%\\
% AWS AI Labs\\
% {\tt\small alongolt@amazon.com}
\quad
Roy Ganz\textsuperscript{2}\footnotemark[2]%\\
% Technion, Israel\\
% {\tt\small ganz@cs.technion.ac.il}
% ,
~\quad
Oren Nuriel\textsuperscript{1}%\\
% AWS AI Labs\\
% {\tt\small onuriel@amazon.com}
\\
Royee Tichauer\textsuperscript{1}%\\
% AWS AI Labs\\
% {\tt\small royeet@amazon.com}
\quad
Shai Mazor\textsuperscript{1}%\\
% AWS AI Labs\\
% {\tt\small smazor@amazon.com}
\quad
Ron Litman\textsuperscript{1}%\\
% AWS AI Labs\\
% {\tt\small litmanr@amazon.com}
\vspace{0.15cm} \\
{\small
\textsuperscript{1}AWS AI Labs \quad \textsuperscript{2}Technion, Israel}
\vspace{-0.3cm}
}

\maketitle
\ificcvfinal\thispagestyle{empty}\fi

\newcommand{\AlgoName}{CLIPTER }
\newcommand{\AlgoNameNoSpace}{CLIPTER}

%%%%%%%%% ABSTRACT
\begin{abstract}
    Reading text in real-world scenarios often requires understanding the context surrounding it, especially when dealing with poor-quality text. However, current scene text recognizers are unaware of the bigger picture as they operate on cropped text images. In this study, we harness the representative capabilities of modern vision-language models, such as CLIP, to provide scene-level information to the crop-based recognizer. We achieve this by fusing a rich representation of the entire image, obtained from the vision-language model, with the recognizer word-level features via a gated cross-attention mechanism. This component gradually shifts to the context-enhanced representation, allowing for stable fine-tuning of a pretrained recognizer. We demonstrate the effectiveness of our model-agnostic framework, CLIPTER (CLIP TExt Recognition), on leading text recognition architectures and achieve state-of-the-art results across multiple benchmarks. Furthermore, our analysis highlights improved robustness to out-of-vocabulary words and enhanced generalization in low-data regimes.
\end{abstract} 

%%%%%%%%% BODY TEXT
\section{Introduction}
\label{sec:introduction}

\begin{figure}[t!]
% \vspace{-0.2cm}
    \centering
    \includegraphics[width=0.95\linewidth]{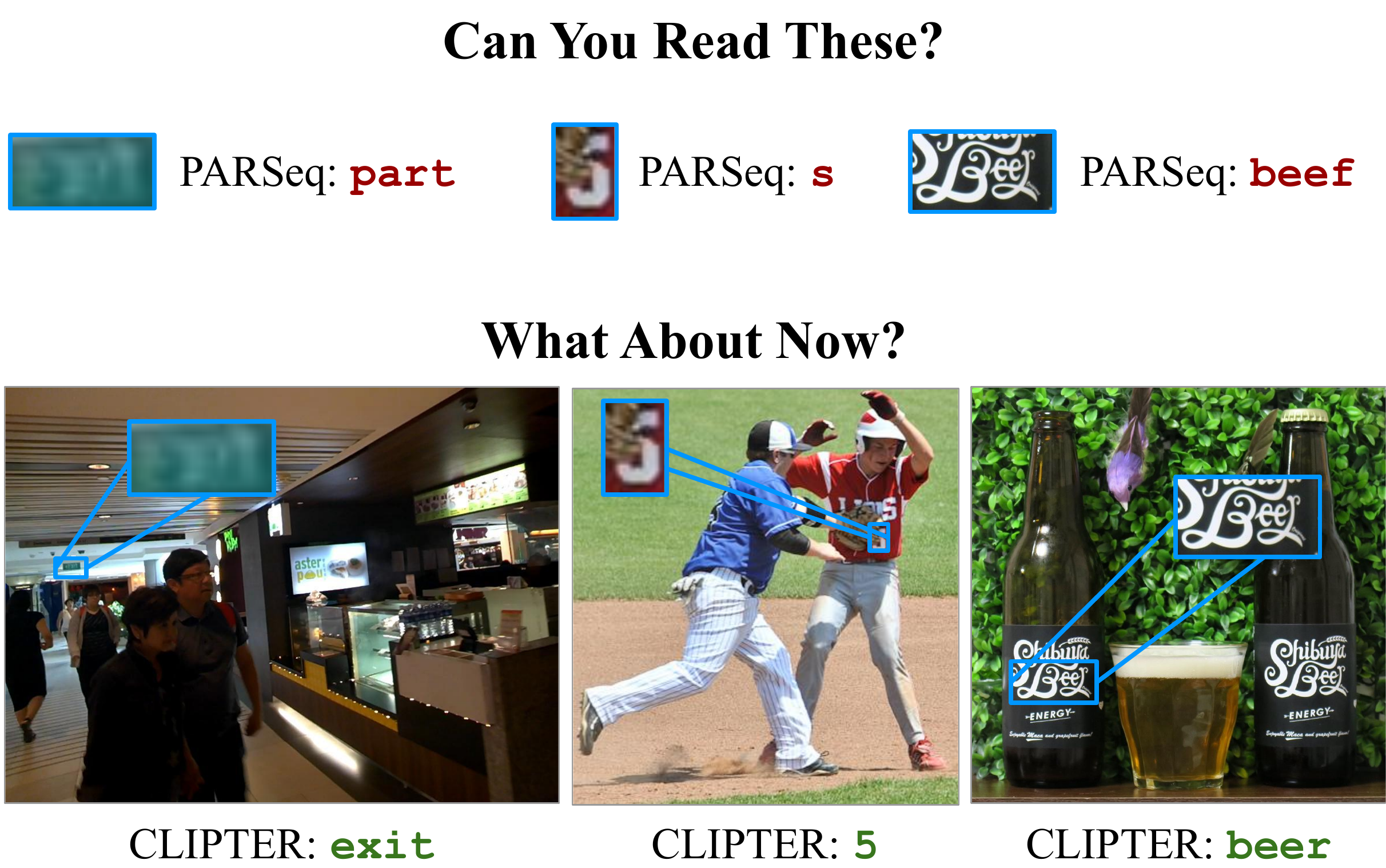}
    \vspace{-0.1cm}
    \caption{\textbf{The Importance of Seeing the Bigger Picture.} 
    Scene context often assists in reading text in real-world scenarios,  and in certain cases, it is even vital.
    Thus, current crop-based text recognizers are inherently limited (Top).
    To address this limitation, our method, CLIPTER, provides the recognizer with scene information (Bottom).
    }
    \label{fig:teaser_examples}
    \vspace{-0.2cm}
\end{figure}

Recognizing text in real-world settings often involves leveraging contextual information from the scene, particularly when dealing with blurry, low-resolution, corrupted, or occluded text, as showcased in \cref{fig:teaser_examples}. 
Conversely, learning-based methods typically detect text in the image and then perform recognition solely on the cropped detected regions, neglecting valuable scene information~\cite{Baek2019clova,litman2020scatter,abinet_fang2021read,baek2021if,aberdam2021sequence,aberdam2022multimodal,vitstr_atienza2021vision,nuriel2021textadain,bautista2022parseq,liu2022perceiving,conclr_zhang2022context,zheng2022pushing,na2022multi}.
As a result, the practice of operating on cropped text images is inherently suboptimal.

\begin{figure}[t!]
\vspace{-0.3cm}
    \centering
    \includegraphics[width=0.9\linewidth]{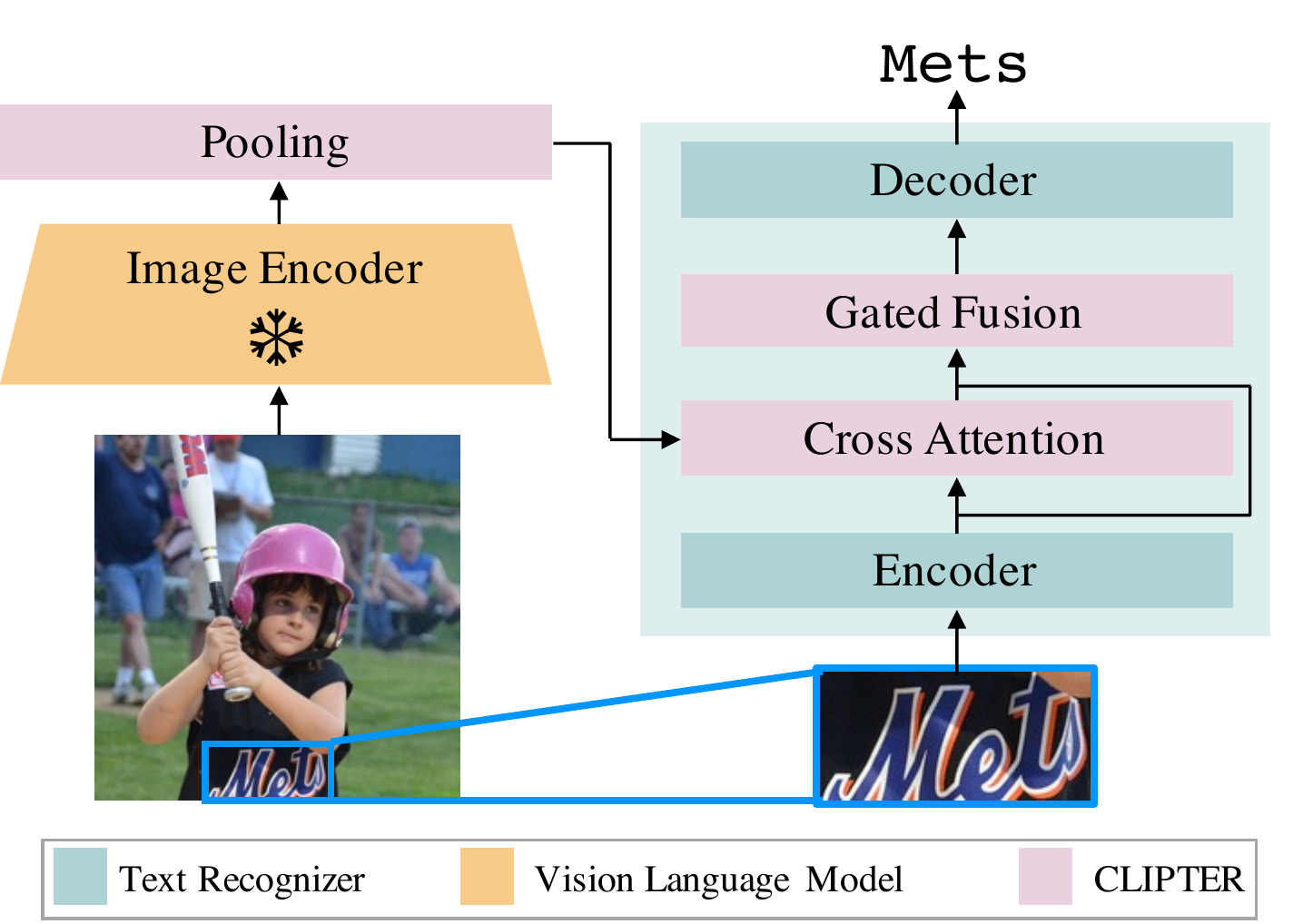}
    \vspace{-0.2cm}
    \caption{
    \textbf{CLIPTER -- Incorporating Scene Context into Text Recognizers.} 
    Our novel approach employs a frozen vision-language model, such as CLIP, to extract rich features of the entire scene image.
    These features are then fused with the crop-level features using our gated cross-attention mechanism, which gradually shifts the pretrained recognizer to the context-enriched features.
    }
    \label{fig:teaser_highlevel}
    \vspace{-0.5cm}
\end{figure}

To overcome this limitation, we explore the use of vision-language models. These models, pretrained on a vast corpus of image-caption pairs, exhibit powerful representation capabilities and can be used for numerous downstream tasks~\cite{radford2021learning,wang2022unifying,minderer2022simple,biten2022latr,alayrac2022flamingo,li2022blip,wang2021simvlm,li2021align}.
Unlike models trained only on visual data, such as MAE~\cite{he2022masked}, vision-language models are also supervised by the corresponding textual description. 
This description brings focus to the crucial details in the scene, which in turn can assist in reading poor-quality text, as we show later. Moreover, the image caption can even contain actual text words in the image, such as business logos and street names, due to their necessity for describing the scene. Hence, leveraging vision-language models can facilitate in recognizing such words, which are typically unique, categorized as out-of-vocabulary, and therefore pose greater difficulty to text recognition models~\cite{wan2020vocabulary}.

In this work, we introduce \AlgoName (CLIP TExt Recognition), a general framework  for integrating image-level knowledge into crop-based text recognizers. To this end, our method first extracts a rich visual representation of the entire image using a vision-language image encoder. As depicted in \cref{fig:teaser_highlevel}, this representation is then merged with the word-level features of the cropped text instance using a cross-attention-based operation.
Additionally, we incorporate a gating mechanism, which gradually shifts between the word-level features and the merged representations during training. This mechanism provides a more stable training process and enables the adaptation of pre-existing models, including those pretrained on synthetic data. 
As a result, \AlgoName can effectively enhance any pretrained recognizer with scene context awareness.

We design our method as a versatile framework consisting of modular blocks of varying sizes that can support various text recognition architectures and adapt to diverse computational constraints.
In particular, we explore a range of vision and vision-language image encoders, pooling operators, light-to-heavy fusion schemes, and different integration points between word-level and image-level representations.
This integration point is critical and highly dependent on the underlying architecture, and therefore we study two types of integration point: early fusion within the vision model, which considers the image representation as additional visual content, and late fusion at the decoding stage, which utilizes the image features as supplementary contextual information to condition the prediction on.

Throughout extensive experimentation on twelve highly-diverse datasets, our method exhibits consistent improvements on top of various leading text recognition methods, such as TRBA~\cite{Baek2019clova}, ABINet~\cite{abinet_fang2021read}, and PARSeq~\cite{bautista2022parseq}. In particular, implementing CLIPTER on PARSeq achieves state-of-the-art (SoTA) results on all benchmarks, including dense text and challenging street-view images.
Further in-depth analysis reveals that incorporating CLIPTER improves robustness to out-of-vocabulary words and enhances generalization capability in low-data regimes. 

To account for all the computations involved in adding CLIPTER to an existing recognizer, we perform an end-to-end evaluation, in which we cascade the recognizer after an existing text detector. This setting not only reveals performance gains over two-stage and end-to-end approaches, but also demonstrate a marginal impact in the overall latency. 
Finally, through a comprehensive ablation study, we develop a recipe for integrating CLIPTER in other text recognition architectures, including future ones.

To summarize, our main contributions are:
\begin{itemize}[topsep=0.1cm]% [nolistsep, leftmargin=*]
\setlength\itemsep{-0.03cm}
    \item Introducing CLIPTER, a framework for enhancing text recognition performance by incorporating scene context through the use of vision-language models.
    \item The design of a computationally efficient and flexible framework that can be incorporated with various existing text recognition architectures.
    \item Demonstrating consistent improvements of leading text recognizers on diverse datasets, achieving state-of-the-art results, and enhancing robustness to out-of-vocabulary and generalization in low-data regimes.
\end{itemize}

%-------------------------------------------------------------------------

\section{Related Work}

\paragraph{\textbf{Scene text recognition.}} 
Significant progress has been made in word-level scene text recognition in recent years~\cite{Baek2019clova,litman2020scatter,qiao2020seed,baek2021if, nuriel2021textadain,wang2021two,srn_yu2020towards,conclr_zhang2022context,slossberg2020calibration,bhunia2021towards}, largely due to the adoption of transformer-based models~\cite{vitstr_atienza2021vision,bautista2022parseq,abinet_fang2021read,na2022multi} and the exploitation of unlabeled data~\cite{seqclr_aberdam2021sequence,liu2022perceiving,lyu2022maskocr,aberdam2022multimodal,zheng2022pushing,luo2022siman}.
%Notable examples 
Recent SoTA method includes ViTSTR~\cite{atienza2021vision} and MaskOCR~\cite{lyu2022maskocr}, which propose simple ViT-based architectures to improve vision extraction, and ABINet~\cite{abinet_fang2021read} and SRN~\cite{srn_yu2020towards}, which incorporate linguistic knowledge through transformer-based language modalities to refine vision model predictions. Additionally, SeqCLR~\cite{seqclr_aberdam2021sequence}, CCR~\cite{zheng2022pushing}, Persec~\cite{liu2022perceiving} and SemiMTR~\cite{aberdam2022multimodal} use contrastive learning and consistency regularization to learn from unlabeled data.
Nevertheless, all these methods suffer from a lack of scene-level context, as they operate on cropped text images. Consequently, in challenging cases of corrupted text, these models resort to predict the most likely word from their training vocabulary~\cite{wan2020vocabulary,garcia2022out}. We address this limitation by enriching the recognizer with scene-level information.

It should be noted that while there is an alternative end-to-end approach called text spotting~\cite{liao2020mask,kittenplon2022towards,zhu2020deformable,zhu2020deformable,xue2022language,huang2022swintextspotter}, which allows the text decoder to access the entire image when decoding the text, our work focuses on improving the cascaded pipeline of separate detection and recognition steps. The cascaded pipeline, which is widely used and studied, offers advantages such as modularity, task decoupling, invariance to scale and rotation~\cite{ronen2022glass}, and efficiency in using synthetic data~\cite{Baek2019clova}.

\vspace{-0.25cm}
\paragraph{\textbf{Vision-language models.}} 
Vision-language models trained on a large number of image-text pairs provide effective representations for various tasks~\cite{zhang2021vinvl,radford2021learning,li2021align,wang2021simvlm,li2022blip,hu2022scaling, wang2022git,chen2022pali, wang2022unifying}. 
Among the pioneers in this area, CLIP~\cite{radford2021learning} used contrastive learning to train image and text encoders to produce aligned representations of image-caption pairs.
BLIP~\cite{li2022blip} proposed a filtering mechanism to handle noisy image-caption pairs, while GIT~\cite{wang2022git} simplified the architecture to only one image encoder and one text decoder. 
Inspired by their potential, we aim to utilize them for scene text recognition.

\section{Methodology}
\label{sec:methodology}

\begin{figure*}[htp!]
    \centering
    \includegraphics[width=0.85\textwidth]{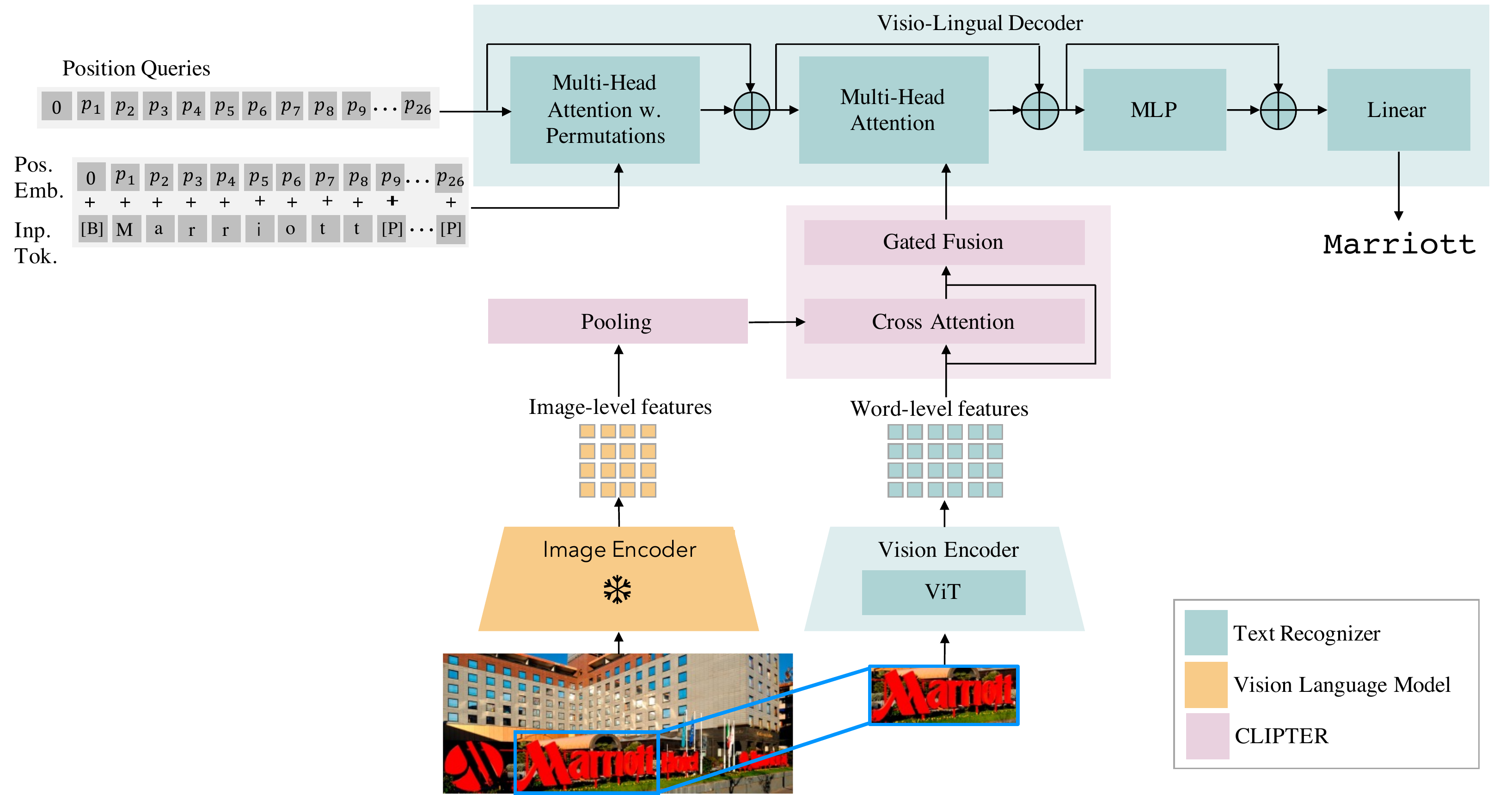}
    \caption{\textbf{An Overview of \AlgoName Integrated into PARSeq~\cite{bautista2022parseq}.} Our framework introduces four building blocks: (1) A pretrained \textit{Image Encoder} used to extract high-level representations from the entire image. (2) These representations are then fed to the \textit{Image Feature Pooling} layer that can pool spatial dimensions, balancing latency and performance. (3) Upon obtaining features, an \textit{Integration Point} (early or late) is chosen to incorporate this information, which can change for different architectures. (4) Lastly, the \textit{Fusion Mechanism} composed of a gated cross-attention mechanism, merges the two streams of information allowing the recognizer to reason over both.
    }
    \label{fig:main_figure}
    \vspace{-0.5cm}
\end{figure*}

Our method proposes a model-agnostic and easy-to-implement framework, which is designed to compensate for the lack of scene context in crop-based text recognizers. In this section, we detail the building blocks, describe the training procedure, discuss the running time, and present several recognition models on which we apply our method.

\subsection{Building Blocks}
\label{subsec:building_blocks}

Our algorithm consists of four elements. First, the image encoder generating the image-level features. %Next, an optional pooling operation can be applied to these features to reduce memory and latency.
Next, an optional pooling operation on the obtained features can reduce memory and latency.
In the third stage, an integration point is determined to incorporate these features into the target text recognizer. Finally, a fusion mechanism merges the image- and word-level representations. 
The algorithm pseudocode is provided in \cref{app:pseudocode} and its implementation on PARSeq architecture~\cite{bautista2022parseq} is illustrated in \cref{fig:main_figure}.

\paragraph{Image Encoder}
The image encoder aims to complement the recognizer word-level information with scene context-aware representations. We explore several powerful encoders, which can be divided into two categories:
\begin{itemize} % [nolistsep, leftmargin=*]
    \item \emph{Vision-based models} such as ViT~\cite{vit}, MAE~\cite{he2022masked} and DiNO~\cite{caron2021emerging}, which are pretrained solely on images and encompass the image visual content, including the class and position information of its objects.
    \item \emph{Vision-language-based models} such as CLIP~\cite{radford2021learning}, BLIP~\cite{li2022blip} and GiT~\cite{wang2022git}, which are pretrained
    on a massive and highly-diversified dataset of images and their textual descriptions.
    These descriptions focus the representations on the crucial details in the scene, and as shown in \cref{sec:ablation}, lead to better performance.
\end{itemize}

Our study focuses on transformer-based~\cite{vit} vision encoders, where the number of output representations corresponds to the number of image patches, denoted as $HW$, plus an additional representation for the special token \texttt{[class]}. To maintain its generalization ability and prevent a substantial increase in training runtime and memory, we keep the image encoder frozen during training.

\paragraph{Image Feature Pooling}
The size of the image features affects training and inference times, as they are integrated with word-level representations using a cross-attention-based mechanism. The computational complexity of this operation is $\mathcal{O}(N_{\text{global}} \cdot N_{\text{local}} \cdot d)$, where $N_{\text{global}}$ and $N_{\text{local}}$ denote the corresponding sequence lengths of the image-level and word-level representations, and $d$ is their dimension. 

To optimize the balance between performance and latency, a pooling component is introduced. This layer preserves the first image-level representation corresponding to the \texttt{[class]} token and applies 2D average pooling to the remaining per-patch representations.
As a result, the output feature size becomes $\rmF^{\text{global}} \in \R^{(1+HW/k^2) \times d}$, where $k$ denotes the pooling kernel.
In \cref{sec:ablation}, we empirically study the trade-off between computational cost and granularity level in the choice of $k$. Surprisingly, our findings reveal that even using only the first representation of CLIP (marked as $k=\infty$) can still improve performance.

\paragraph{Integration Point}
Another decision is where to inject the global, image-level features within the recognition model. Since recognition architectures differ significantly, we presume there is no fixed integration point and thus explore several options for each recognizer.  
In \cref{subsec:recognition_models}, we describe the studied recognition schemes and define optional integration points for each. 
However, in general, these integration points can be divided into two main categories: 
\begin{itemize} % [nolistsep] 
    \item \emph{Early fusion} -- The integration is performed in the vision encoder and targets the visual features extracted by a convolutional, or transformer-based backbone. This approach views the global features as visual content and therefore merges them with the visual features of the crop. In some architectures, there are several options for an early integration point. 
    
    \item \emph{Late fusion} -- The integration is performed in the linear, attention-based or transformer-based decoder. This fusion can be seen as conditional decoding, in which the characters are decoded given the image state. In autoregressive decoders, such integration leads to a significant increase in inference time, repeating the cross-attention operation in each decoding step.
\end{itemize}
Note that early and late fusion approaches have been studied in the literature~\cite{li2019visualbert, joze2020mmtm,appalaraju2021docformer,li2021align}, although not in our context of merging local and global information in text recognition.

\paragraph{Fusion Mechanism}
The role of this component is fusing the image-level and word-level features, $\rmF^{\text{global}}$ and $\rmF^{\text{local}}$, correspondingly. To this end, we first linearly project the global features to the dimension of the local representations, $d$. Then, we choose between two attention-based schemes:
\begin{itemize} % [nolistsep] 
    \item \emph{Multi-head cross-attention (MH-CA)} -- The nowadays natural approach of combining two data streams of different resources~\cite{attention_is_all_vaswani2017attention,devlin2018bert}. In our case, queries are local features, and global features are the keys and values:
    \begin{equation*}
        \rmF^{\text{mixed}} = \operatorname{MH-CA}(Q{=}\rmF^{\text{local}}, K{=}\rmF^{\text{global}}, V{=}\rmF^{\text{global}}).
    \end{equation*}
    We examine several, compact to heavy, models with a different number of attention heads, hidden layers, hidden sizes, and intermediate sizes.
    
    \item \emph{Gated attention} -- A lightweight alternative that applies a gated mechanism between global and local features. Such a model cannot handle different lengths of representation sequences. Therefore, it can be utilized only if there is a single image-level representation after the pooling ($k=\infty$). In this case, for each local representation $\rvf^{\text{local}}_i \in \R^{d}$ we independently apply: 
    \begin{align}
        \rvg & = \softmax( \rmW \left[ \rvf^{\text{local}}_i; \rvf^{\text{global}} \right] ) \\
        \rvf^{\text{mixed}}_i & = \rvg \circ \rvf^{\text{local}}_i + (\rvone - \rvg) \circ \rvf^{\text{global}},
    \end{align}
    where $\rmW \in \R^{d \times 2d}$ is a weight matrix, and $\circ$ is an elementwise Hadamard product. 
\end{itemize}

Our training starts with a pretrained baseline model, which we fine-tune to become context-aware. 
To enhance the stability of this process, we implement a $\tanh$-gating mechanism inspired by \cite{hochreiter1997long,alayrac2022flamingo}. This mechanism preserves the forward-pass intact during initialization and gradually transitions between the original word-level features and the fused representation throughout training:
\begin{equation}
    \rmF^{\text{fused}} = (1-\tanh(\alpha)) \rmF^{\text{local}} + \tanh(\alpha)  \rmF^{\text{mixed}},
\end{equation}
where $\alpha$ is a learnable scalar initialized at 0.

\subsection{Training Protocol}
\label{subsec:training_protocol}

The CLIPTER framework is a versatile solution that can be used for various text recognition schemes. Instead of adjusting the training parameters for each recognition platform, we employ a pretrained baseline model to initialize all the original parts. Then, we only fine-tune the text recognizer and the fusion mechanism, with hyperparameters that are agnostic to the chosen recognition architecture. This approach reduces the training time and allows for the utilization of synthetic data in training the baseline model.

As the image encoder is fixed in our approach, we can save training time and memory by preparing the dataset in advance. This involves pre-computing the image-level representations by passing all training images through the image encoder. The resulting dataset can be used for training different recognition architectures. To further reduce data loading latency, we cache the image-level representations during the first epoch of training. These measures lead to a minor increase in the training time of each iteration, less than $10\%$ when using a reasonable fusion mechanism.

\input{Tables/sota_performance}

\subsection{Studied Recognition Models}
\label{subsec:recognition_models}

Here, we detail at a high-level the text recognition models we examine, as well as suggested integration points.
\begin{itemize} [topsep=0.1cm]% [nolistsep, leftmargin=*]
\setlength\itemsep{-0.03cm}
    \item \emph{TRBA} -- A general architecture~\cite{Baek2019clova}, which comprises four components: 
    (i) transformation for normalization of the input image, (ii) a ResNet-based visual feature extractor, (iii) a Bi-LSTM-based contextual block, and (iv) an attention decoder.
    We explore three integration points: \emph{visual} and \emph{contextual}, corresponding to early fusion after stages (ii) and (iii), and \emph{decoder}, of fusing representations in the prediction block of stage (iv).

    \item \emph{ViT-STR} -- This scheme consists of a single stage, and hence, the integration point is at the output of the ViT.

    \item \emph{ABINet} -- A multimodal scheme~\cite{abinet_fang2021read}, comprising three components: 
     (i) a vision model with a ResNet backbone network and transformer unit, (ii) a language model that refines the output vision embeddings, and (iii) a fusion model that combines the output of the vision and language models for final prediction.
    We investigate two early integration points within the vision model (stage i): \emph{visual} after the ResNet backbone, and \emph{contextual} after the transformer unit, as well as a late fusion: \emph{decoder} within the fusion model (stage iii). 

    \item \emph{PARSeq} -- A transformer-based architecture~\cite{bautista2022parseq}, depicted in \cref{fig:main_figure}. Here we define an early integration: \emph{visual} after the ViT model, and late integration: \emph{decoder} after the cross attention block.
\end{itemize}

\section{Experiments}

We hereby present a comprehensive evaluation of our approach in combination with state-of-the-art text recognizers on twelve diverse scene text recognition benchmarks. Our study showcases the broad applicability of our proposed method, as it consistently outperforms existing approaches across all datasets and architectures. 
Moreover, we conduct an in-depth analysis of our method's generalization capability on images containing out-of-vocabulary words, revealing a significant improvement in performance. Finally, we demonstrate that our approach surpasses current methods in scenarios with limited amounts of labeled training data.

\vspace{0.1cm} \noindent \textbf{Datasets.}
We follow the pre-processing of \cite{whatif_baek2021if,aberdam2022multimodal} and similarly perform experiments using only real data. We demonstrate our results on twelve scene text benchmarks: ICDAR 2013~\cite{karatzas2013icdar}, ICDAR 2015~\cite{karatzas2015icdar}, ArT~\cite{chng2019icdar2019}, SVT~\cite{wang2011end}, LSVT~\cite{lsvt_sun2019chinese}, COCO-Text~\cite{veit2016coco}, RCTW~\cite{shi2017icdar2017}, Uber~\cite{zhang2017uber}, ReCTS~\cite{zhang2019icdar}, MLT19~\cite{nayef2019icdar2019}, HierText~\cite{long2022towards} and TextOCR~\cite{singh2021textocr}. Both TextOCR and HierText are particularly large datasets, rich with text and containing $30$ and $100$ words on average per image, respectively. In total, our test set contains over 200k words, 20 times larger than similar works~\cite{litman2020scatter,Baek2019clova, abinet_fang2021read,seqclr_aberdam2021sequence,nuriel2021textadain}. Dataset characteristics and train/test splits are provided in \cref{app:datasets}. We evaluate performance using word-level accuracy and normal/weighted averages across datasets.

\vspace{0.1cm} \noindent \textbf{Implementations Details.}
\label{sec:implementations_details}
We adopt the codebase of PARSeq\footnote{\notsotiny \url{https://github.com/baudm/parseq}}~\cite{bautista2022parseq} and SemiMTR\footnote{\notsotiny \url{https://github.com/amazon-science/semimtr-text-recognition}}~\cite{aberdam2022multimodal} to establish our baselines. Our experiments are conducted on 4 Tesla V100 GPUs, 16GB memory, using PyTorch.
We closely follow the configuration parameters of the baseline models, as detailed in \cref{app:implementation_details}.
We use a 36-character set (10 digits and 26 letters) in most experiments, except for using the full 94-character set in \cref{subsec:oov}. 
We introduce a range of cross-attention models, including \emph{gated attention} and three multi-head cross-attention (MH-CA) types: \emph{tiny}, \emph{mini} and \emph{small}, containing 328K, 923K, 5.3M and 18.1M parameters, respectively, and are further elaborated in \cref{app:implementation_details}.

\begin{figure*}[htp!]
    \centering
    \includegraphics[width=\textwidth]{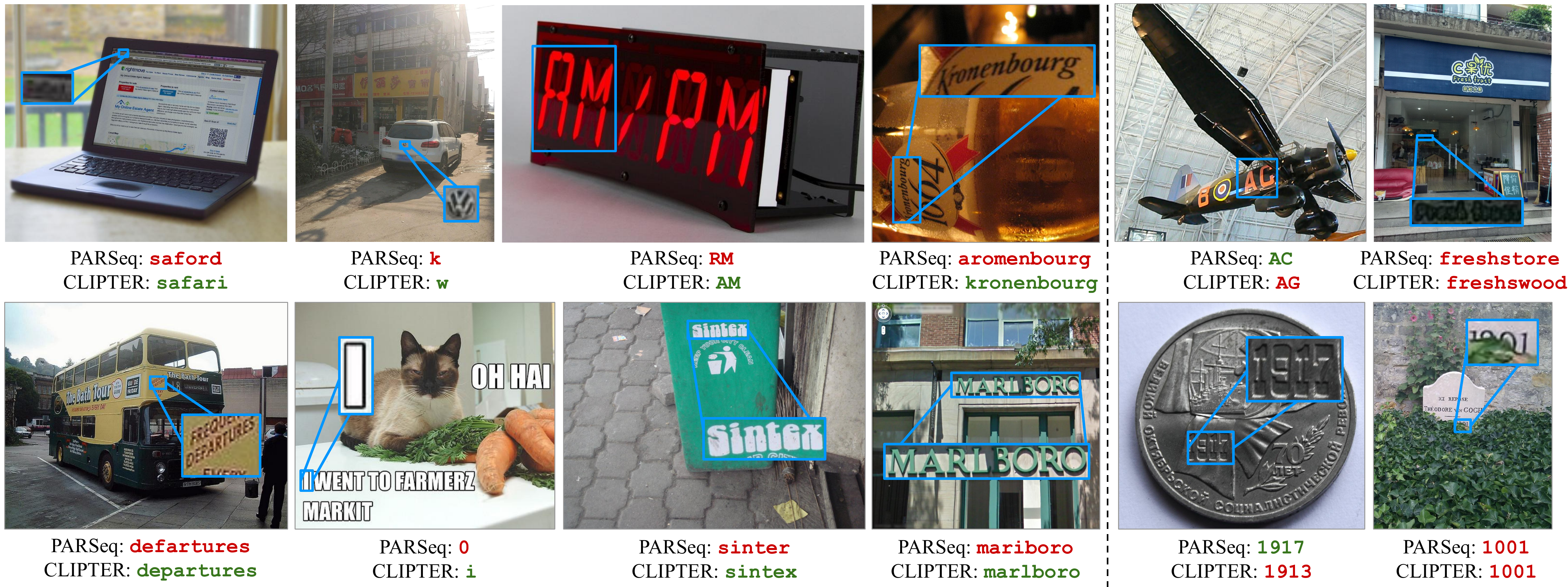}
    \vspace{-0.4cm}
    \caption{\textbf{Qualitative Examples.} 
    The eight images on the left depict failure cases of the baseline model PARSeq~\cite{bautista2022parseq}, that become success cases when incorporating \AlgoName.
    On the right, we present examples where our method produces incorrect predictions and where both models fail to correctly decode the text.
    }
    \label{fig:results}
    \vspace{-0.4cm}
\end{figure*}

\subsection{Improving State-of-the-Art Recognizers}
In \cref{table:sota}, we examine the impact of CLIPTER on leading scene text recognition methods across 12 diverse benchmarks. Despite the diversity in architectures, including CNN-based and transformer-based visual encoders, and autoregressive and parallel decoders, we demonstrate that CLIPTER significantly improves performance for all tested methods. In particular, the improvements in accuracy weighted average are $+0.9\%$ in TRBA, $+1.4\%$ in ViT-STR, $+1.7\%$ in ABINet-VIS, and $+0.4\%$ in ABINet. Moreover, we establish new SoTA results by integrating CLIPTER with PARSeq, the current top-performing text recognizer, increasing its accuracy by $+0.8\%$ across all datasets. In \cref{fig:results}, we present qualitative results of successful and unsuccessful cases, with additional examples in \cref{app:qualitative}.

Breaking down the evaluation datasets reveals that our method is especially beneficial for street-view datasets, namely Uber, SVT and LSVT, decreasing the relative error of PARSeq by almost 10\%. Uber-Text data~\cite{zhang2017uber} is predominantly comprised of street names and business logos, presenting multiple challenging text instances that are blurry, occluded, or of low-resolution. In such cases, the surrounding context plays a vital role, even for human perception.
Furthermore, our method exhibits improvements on text-rich datasets, TextOCR and HierText, with an average of 30 and 100 words per image~\cite{long2022towards}, compared to less than 10 words in other datasets.
These results demonstrate that leveraging image-level information can be advantageous even for text-dense images and documents, indicating the potential of vision-language models to reason from text in images. See further analysis in \cref{app:dense_documents}.

As expected, we observe that a one-size-fits-all approach is not feasible due to the vast differences in the architectures of text recognizers. Therefore, for each recognition scheme, we present the best results achieved using CLIPTER and leave the discussion of design choices to \cref{sec:ablation}.
More precisely, in \cref{table:sota}, we denote the integration point in subscript, use the MH-CA mini for the fusion mechanism, and the image encoder is BLIP (with pooling of $k=5$) for ABINet and CLIP ($k=\infty$) for the others.

\subsection{Performance on Out-Of-Vocabulary}
\label{subsec:oov}
\input{Tables/oov}
\begin{figure}[t!]
    \centering
    \vspace{-0.1cm}
    \includegraphics[width=0.7\linewidth]{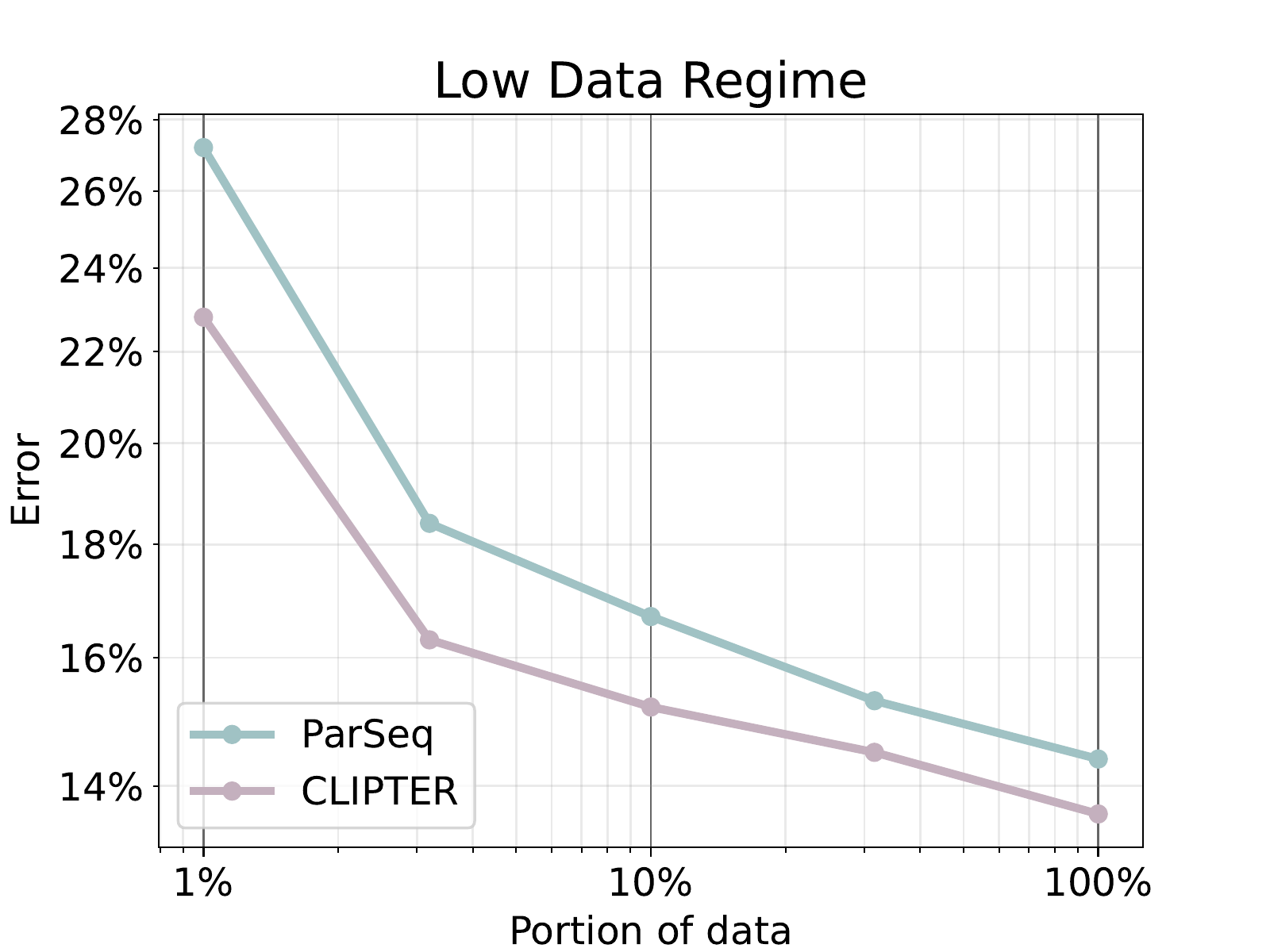}
    \vspace{-0.2cm}
    \caption{\textbf{Word Error Rate Versus Data Portion in Log-Log Scale.} Our method, trained on 40\% of the data, reaches the baseline performance when trained on the entire data. }
    \label{fig:low_data}
    \vspace{-0.4cm}
\end{figure}

Motivated by the improved results on street-view images, we %proceed to 
examine our method on out-of-vocabulary (OOV) text instances -- words that do not appear in the training sets. These %text instances 
are often crucial for understanding the scene, as they can contain prices, names, dates, phone numbers, emails, and URLs. 
However, as shown in~\cite{wan2020vocabulary,garcia2022out}, current methods over-rely on their train vocabulary, especially in low-quality or distorted text. 
To test if scene context can assist in these cases, we utilize the newly proposed OOV benchmark~\cite{garcia2022out}. As demonstrated in \cref{tab:oov}, integrating CLIPTER into PARSeq, not only improves accuracy by 1.52\% on general words, but is even more significant in OOV words, presenting an improvement of 2.48\%. The robustness to OOV words is yet another reason to harness the knowledge of massively pretrained vision-language models.

\subsection{Performance in Low Data Regime}
To further probe the benefits of our method, we evaluate its performance in the low data regimes of 1\%, 5\%, 10\%, and 25\% of the training data.
As shown in \cref{fig:low_data}, \AlgoName leverages the generalization power of the vision-language model and thus becomes even more effective and beneficial when the amount of training data decreases.
In particular, training \AlgoName on 10\% of the data leads to similar results as the baseline trained on 25\%. Likewise, when training \AlgoName on 40\%, it achieves the performance of the baseline trained on 100\% of the data. Similar trends appear with TRBA and ViTSTR, as shown in \cref{app:low_data}.

\section{End-to-End Latency and Performance}
\label{sec:e2e_lightweight}
\input{Tables/end2end}
\input{Tables/lightweight}

To accurately measure the impact of our method on latency, we aim to account for all its components and recognize that the encoding of the entire image is computed only once, regardless of the number of words it contains. Therefore, instead of calculating latency of standalone recognition on a single cropped text image, we construct an end-to-end evaluation that better simulates real-world latency. For this purpose, we employ the text detector of GLASS~\cite{ronen2022glass} and cascade it with PARSeq, both with and without CLIPTER. Here, we focus on a lightweight version of our method that consists of $\text{CLIP}_\text{base}$ image encoder and gated attention fusion mechanism. Our results, as shown in \cref{tab:end2end}, indicate that our method adds only 8\% to the overall latency (+12 ms per image) while delivering superior performance that outperforms both two-stage pipelines and existing E2E text spotting methods. For completeness, we present the recognition results of the lightweight version in \cref{tab:lightweight}, demonstrating nearly optimal performance, and offer further implementation details in \cref{app:latency}.

\section{Ablation Studies}
\label{sec:ablation}
Here, we study the relative effect of each component in our scheme, including choice of image encoder, pooling kernel size, integration point and fusion mechanism. Throughout our analysis, we discuss the performance-latency tradeoff and provide general recommendations for integrating CLIPTER in other text recognition methods. 

\input{Tables/image_encoders}
\vspace{0.1cm}\noindent{\textbf{The Choice of the Image Encoder.}}
The first part of \cref{table:abla_image_encoders} exhibits the performance of PARSeq with CLIPTER, when leveraging the vision-based image encoders of DiNO, ViT-MAE and OWL-ViT, and when using the vision-language models of CLIP, BLIP and GIT.
As shown, the best performance is achieved with the latter models. These models were pretrained not only on images, but also on their textual descriptions, leading to more informative and effective representations. Interestingly, the compact representation of CLIP leads to the best results in PARSeq. This, however, is not the case for all recognizers. For example, ABINet benefits more from the larger representation of BLIP. In general, though, the go-to method is still the single representation of CLIP, as it yields nearly the best performance and demonstrates low computation cost.

In the second part of \cref{table:abla_image_encoders}, we examine the effect of the pooling kernel $k$. To this end, we apply CLIPTER on TRBA with a fixed image encoder, BLIP, and only change the kernel size. 
As shown, too aggressive pooling ($k=\infty$) deteriorates representation quality. However, besides this extreme, varying $k$ does not impact performance significantly but can lead to severe consequences on running times. Thus, our recommendation is to use a relatively coarse representation of the scene, which performs decently well.

\input{Tables/integration_point}
\vspace{0.1cm}\noindent{\textbf{Integration Point.}}
In \cref{tab:intergration_point}, we evaluate the effect of the integration point on our studied architectures, considering the three types defined in \cref{sec:methodology}: \emph{vision} and \emph{contextual} in early fusion, and \emph{decoder} in late fusion. As shown, TRBA and PARSeq are less sensitive to this decision; whereas ABINet benefits from late fusion and ABINet-Vis from early, vision fusion.
Note, however, that the runtime complexity of late fusion increases dramatically for autoregressive decoders, as in PARSeq and TRBA, but not for parallel decoders, as in ABINet.
As expected, the significant differences between the text recognition architectures imply that the decision of the integration point is not clear-cut and thus, integrating CLIPTER in new architectures require an empirical search to locate the optimal fuse point.

\input{Tables/ablations}
\vspace{0.1cm}\noindent{\textbf{Fusion Mechanism.}}
We examine the effect of the fusion model capacity, considering a compact gated attention scheme to more complex multi-head attention modules.
Since gated-attention can be applied only for a single vector, we examine it with CLIP image encoder ($k=\infty$). For the other alternatives, we also consider the spatial representations of BLIP with $k=5$. As shown in \cref{table:fusion}, more complex schemes usually lead to better results, but at the cost of additional computational load, represented by the number of FLOPS. We find the gated-attention and MH-CA mini as balancing points between quality and runtime.

\section{Conclusions}

We introduced CLIPTER, a novel approach to enrich crop-based text recognizers with scene knowledge, by utilizing vision-language models. 
Our versatile framework is composed of modular blocks, which enable the fine-tuning of various pretrained text recognition architectures.
Our extensive experiments on diverse benchmarks demonstrate that incorporating CLIPTER into existing approaches consistently enhances their performances, demonstrating better generalization and robustness to out-of-vocabulary words.
Moreover, our end-to-end evaluation revealed a marginal increase in the overall latency, while presenting improved results, even compared to text spotting methods. 
Finally, through a comprehensive ablation study, we provide guidelines for implementing our method on future recognizers, paving the way for further advancements in this area.

\clearpage

%%%%%%%%% REFERENCES

{\small
\bibliographystyle{ieee_fullname}
\bibliography{egbib}
}

\clearpage

\appendix

\input{appendix_content}

\end{document}

%% file: math_commands.tex
\usepackage{amsmath,amsfonts,bm}

\def\eqref#1{equation~\ref{#1}}
\def\1{\bm{1}}

\def\rvf{{\mathbf{f}}}
\def\rvg{{\mathbf{g}}}

\def\rvone{{\mathbf{1}}}

\def\rmF{{\mathbf{F}}}

\def\rmW{{\mathbf{W}}}

\DeclareMathAlphabet{\mathsfit}{\encodingdefault}{\sfdefault}{m}{sl}
\SetMathAlphabet{\mathsfit}{bold}{\encodingdefault}{\sfdefault}{bx}{n}

\newcommand{\R}{\mathbb{R}}

\newcommand{\softmax}{\mathrm{softmax}}

%% file: Tables/sota_performance.tex
\begin{table*}
\normalsize
\begin{center}
\footnotesize
\bgroup
\def\arraystretch{1.1}
        \resizebox{1\textwidth}{!}{%
        \begin{tabular}{lcccccccccccccc}
        \toprule
% \begin{table*}[!t]
% 	\centering
% 	\resizebox{1\textwidth}{!}{%
% 	\begin{tabular}	{@{\extracolsep{1pt}}lcccccccccccccc}
%         \toprule	 	
        \multirow{2}{*}{\textbf{Method}} &  \textbf{SVT} & \textbf{IC13} & \textbf{IC15} & \textbf{COCO} & \textbf{RCTW} & \textbf{Uber} & \textbf{ArT} & \textbf{LSVT} & \textbf{RECTS} & \textbf{MLT19} & \textbf{TextOCR} & \textbf{HierText} & \textbf{Average} & \textbf{Weighted} \\
        & 647 & 757 & 2,077 & 5,716 & 962 & 49,561 & 3,677 & 3,911 & 2,219 & 4,100 & 70,597 & 75,829 & 220,053 & \textbf{Average} \\
        \midrule
        \rowcolor{LightGray} TRBA \cite{Baek2019clova} & 94.9 & 98.5 & 84.8 & 79.2 & 81.1 & 80.5 & 89.2 & 77.9 & 90.4 & 90.7 & 82.9 & 85.1 & 86.3 & 83.4 \\
        % \cdashline{1-15}
        + $\text{\AlgoName}_{\text{Vision}}$ & 95.4 & 98.8 & 85.3 & 79.3 & 81.3 & 82.0 & 90.2 & 79.4 & 91.1 & 91.1 & 83.9 & 85.8 & 87.0 & 84.3 \\
        $\quad \quad \Delta$ & {\color{OliveGreen}\textbf{+0.5}} & {\color{OliveGreen}\textbf{+0.3}} & {\color{OliveGreen}\textbf{+0.5}} & {\color{OliveGreen}\textbf{+0.1}} & {\color{OliveGreen}\textbf{+0.2}} & {\color{OliveGreen}\textbf{+1.5}} & {\color{OliveGreen}\textbf{+1.0}} & {\color{OliveGreen}\textbf{+1.5}} & {\color{OliveGreen}\textbf{+0.7}} & {\color{OliveGreen}\textbf{+0.4}} & {\color{OliveGreen}\textbf{+1.0}} & {\color{OliveGreen}\textbf{+0.7}} & {\color{OliveGreen}\textbf{+0.7}} & {\color{OliveGreen}\textbf{+0.9}} \\
        \midrule

        \rowcolor{LightGray} ViTSTR-S \cite{vitstr_atienza2021vision} & 92.3 & 97.0 & 81.8 & 77.0 & 72.9 & 77.4 & 86.9 & 73.7 & 88.5 & 89.4 & 80.4 & 83.2 & 83.4 & 80.9 \\
        + $\text{\AlgoName}_{\text{Vision}}$ & 93.4 & 97.1 & 82.3 & 77.7 & 75.3 & 79.6 & 88.2 & 76.0 & 89.5 & 89.9 & 81.8 & 84.0 & 84.6 & 82.3 \\
        $\quad \quad \Delta$ & {\color{OliveGreen}\textbf{+1.1}} & {\color{OliveGreen}\textbf{+0.1}} & {\color{OliveGreen}\textbf{+0.5}} & {\color{OliveGreen}\textbf{+0.7}} & {\color{OliveGreen}\textbf{+2.4}} & {\color{OliveGreen}\textbf{+2.2}} & {\color{OliveGreen}\textbf{+1.3}} & {\color{OliveGreen}\textbf{+2.3}} & {\color{OliveGreen}\textbf{+1.0}} & {\color{OliveGreen}\textbf{+0.5}} & {\color{OliveGreen}\textbf{+1.4}} & {\color{OliveGreen}\textbf{+0.8}} & {\color{OliveGreen}\textbf{+1.2}} & {\color{OliveGreen}\textbf{+1.4}} \\
        \midrule

        \rowcolor{LightGray} ABINet-Vis \cite{abinet_fang2021read} & 88.4 & 97.0 & 80.7 & 75.8 & 72.2 & 75.8 & 85.5 & 72.0 & 87.4 & 89.0 & 78.7 & 83.0 & 82.1 & 79.8 \\
        + $\text{\AlgoName}_{\text{Vision}}$ & 91.8 & 97.0 & 82.0 & 77.1 & 76.1 & 78.1 & 87.4 & 74.9 & 87.8 & 89.4 & 80.6 & 84.4 & 83.9 & 81.5 \\
        $\quad \quad \Delta$ & {\color{OliveGreen}\textbf{+3.4}} & 0.0 & {\color{OliveGreen}\textbf{+1.3}} & {\color{OliveGreen}\textbf{+1.3}} & {\color{OliveGreen}\textbf{+3.9}} & {\color{OliveGreen}\textbf{+2.3}} & {\color{OliveGreen}\textbf{+1.9}} & {\color{OliveGreen}\textbf{+2.9}} & {\color{OliveGreen}\textbf{+0.4}} & {\color{OliveGreen}\textbf{+0.4}} & {\color{OliveGreen}\textbf{+1.9}} & {\color{OliveGreen}\textbf{+1.4}} & {\color{OliveGreen}\textbf{+1.8}} & {\color{OliveGreen}\textbf{+1.7}} \\
        \midrule
        \rowcolor{LightGray} ABINet \cite{abinet_fang2021read} & \textbf{96.6} & 97.6 & 85.1 & 79.4 & 76.7 & 80.8 & 89.2 & 76.6 & 89.4 & 90.2 & 83.1 & 86.6 & 85.9 & 83.9 \\
        + $\text{\AlgoName}_{\text{Decoder}}$  & 96.0 & 98.3 & 85.4 & 79.3 & 78.6 & 82.1 & 89.3 & 77.1 & 89.7 & 90.2 & 83.4 & 86.7 & 86.3 & 84.3 \\
        $\quad \quad \Delta$ &
        {\color{BrickRed}-0.6} & {\color{OliveGreen}\textbf{+0.7}} & {\color{OliveGreen}\textbf{+0.3}} & {\color{BrickRed}-0.1} & {\color{OliveGreen}\textbf{+1.9}} & {\color{OliveGreen}\textbf{+1.3}} & {\color{OliveGreen}\textbf{+0.1}} & {\color{OliveGreen}\textbf{+0.5}} & {\color{OliveGreen}\textbf{+0.3}} & {0} & {\color{OliveGreen}\textbf{+0.3}} & {\color{OliveGreen}\textbf{+0.1}} & {\color{OliveGreen}\textbf{+0.4}} & {\color{OliveGreen}\textbf{+0.4}} \\
        \midrule
        \rowcolor{LightGray} PARSeq \cite{bautista2022parseq} & 96.1 & 98.9 & 85.7 & 80.5 & 81.4 & 83.2 & 91.2 & 80.2 & \textbf{91.8} & 91.5 & 85.2 & 87.4 & 87.8 & 85.6 \\
        + $\text{\AlgoName}_{\text{Vision}}$ & \textbf{96.6} & \textbf{99.1} & \textbf{85.9} & \textbf{81.0} & \textbf{82.1} & \textbf{84.4} & \textbf{91.7} & \textbf{81.8} & \textbf{91.8} & \textbf{91.6} & \textbf{86.0} & \textbf{88.0} & \textbf{88.3} & \textbf{86.4} \\
        $\quad \quad \Delta$ & {\color{OliveGreen}\textbf{+0.5}} & {\color{OliveGreen}\textbf{+0.2}} & {\color{OliveGreen}\textbf{+0.2}} & {\color{OliveGreen}\textbf{+0.5}} & {\color{OliveGreen}\textbf{+0.7}} & {\color{OliveGreen}\textbf{+1.2}} & {\color{OliveGreen}\textbf{+0.5}} & {\color{OliveGreen}\textbf{+1.6}} & 
        0 &
        {\color{OliveGreen}\textbf{+0.1}} & {\color{OliveGreen}\textbf{+0.8}} & {\color{OliveGreen}\textbf{+0.6}} & {\color{OliveGreen}\textbf{+0.5}} & {\color{OliveGreen}\textbf{+0.8}} \\
        \bottomrule
    \end{tabular}
    }
\egroup
\vspace{-0.2cm}
\tiny
\caption{\textbf{Results on the Scene Text Benchmarks.} Scene text recognition accuracy (\%) over twelve public benchmarks. The number of words in each dataset is listed below its name, and we report average and weighted average results.
Integrating CLIPTER into top-performing recognition architectures consistently improves performance. In particular, \AlgoName advances the
state-of-the-art performance of PARSeq~\cite{bautista2022parseq} by +0.5\% and +0.8\% on average and weighted average, respectively.}
\label{table:sota}
\vspace{-0.8cm}
\end{center}
\end{table*}

%% file: Tables/oov.tex
\begin{table}[t]
    \centering
    \small
    \resizebox{0.8\linewidth}{!}{
    \begin{tabular}{lccc}
    \toprule
        \multirow{2}{*}{\textbf{Method}} & \textbf{OOV} & \textbf{IV} & \textbf{Average} \\
        & 25,647 & 91,191 & 116,838 \\
        \midrule
        \rowcolor{LightGray} PARSeq \cite{bautista2022parseq} & 68.97 & 79.74 & 77.38 \\
        + $\text{\AlgoName}_{\text{Vision}}$ & \textbf{71.45} & \textbf{80.99} & \textbf{78.9} \\
        $\quad \quad \Delta$ & {\color{OliveGreen}\textbf{+2.48}} & {\color{OliveGreen}\textbf{+1.25}} & {\color{OliveGreen}\textbf{+1.52}} \\
        \bottomrule
    \end{tabular}
    }
    \vspace{-0.2cm}
    \caption{\textbf{Out-Of-Vocabulary.} %
    Our method not only leads to an improvement over in-vocabulary words, but also to a significant boost on out-of-vocabulary words.
    }
    \vspace{-0.3cm}
    \label{tab:oov}
\end{table}

%% file: Tables/end2end.tex
\begin{table}[t]
\centering
\resizebox{1\linewidth}{!}{
\begin{tabular}{llcccc}
    \toprule
    & \multirow{2}{*}{Method}   & \multicolumn{2}{c}{ICDAR 2015} &  \multicolumn{2}{c}{Total-Text} \\
    \cmidrule(lr){3-4} \cmidrule(lr){5-6}
    & & E2E (G) & FPS &  Word-Spotting (None) & FPS \\ 
    \midrule
    \parbox[t]{1mm}{\multirow{4}{*}{\rotatebox[origin=c]{90}{\textbf{\small E2E}}}}
     % TextDragon~\cite{feng2019textdragon} & 65.2 & - \\
    & ABCNet v2~\cite{liu2021abcnetV2} & 73.0 & 6.5 & 70.4 & - \\
    & Mask TextSpotter v3~\cite{liao2020mask} & 74.2 & 2.6 & - & 2.2 \\
    % Text Perc. & 65.1 & 69.7 \\
    % CRAFTS~\cite{baek2020crafts} & 74.9 & - \\
    & MANGO~\cite{qiao2020mango} & 73.9 & 4.3 & 72.9 & 4.3 \\
    % YAMTS*~\cite{krylov2021yamts} & 74.0 & - \\
    & GLASS~\cite{ronen2022glass} & {76.3} & \textbf{7.75} & {79.9} & \textbf{7.5} \\
    \midrule
    \parbox[t]{1mm}{\multirow{2}{*}{\rotatebox[origin=c]{90}{\textbf{\small 2-STG}}}}
    & GLASS + PARSeq~\cite{bautista2022parseq} & {77.3} & 6.7 & {79.8} & 6.3 \\
    & GLASS + CLIPTER & \textbf{77.4} & 6.2 & \textbf{80.6} & 5.9 \vspace{0.05cm} \\
    \bottomrule
\end{tabular}
}
\vspace{-0.2cm}
\caption{
\textbf{E2E Text Spotting.}
We compare end-to-end (E2E) methods against two-stage (2-STG) pipelines that use GLASS for text detection and PARSeq, with and without CLIPTER, for text recognition. 
Although CLIPTER increases E2E latency by 10 ms per image, it improves SoTA results of GLASS by {\color{OliveGreen}\textbf{+0.9}} on IC15 and {\color{OliveGreen}\textbf{+0.7}} on Total-Text.
% We replaced the recognition head of GLASS (ECCV `22) with PARSeq + CLIPTER and compare it to current SOTA text spotting models. This improves results by {\color{OliveGreen}\textbf{+0.9}} on IC15 and {\color{OliveGreen}\textbf{+0.7}} on Total-Text.
}
\vspace{-0.1cm}
\label{tab:end2end}
\end{table}

%% file: Tables/lightweight.tex
\begin{table}
\begin{center}
\small
% \vspace{-0.1cm}
\resizebox{0.65\linewidth}{!}{%
    \begin{tabular}{lcc}
        \toprule
        \multirow{2}{*}{\textbf{Method}} & \textbf{Average} & \textbf{Weighted} \\
        & 220,053 & \textbf{Average} \\
        \midrule
        \rowcolor{LightGray} TRBA [8] & 86.28 & 83.38 \\
        + $\text{CLIPTER}_{\text{Lightweight}}$ & {\color{OliveGreen}\textbf{+0.43}} & {\color{OliveGreen}\textbf{+0.71}} \\
        \midrule
        \rowcolor{LightGray} ViTSTR-S [6] & 83.37 & 80.89 \\
        + $\text{CLIPTER}_{\text{Lightweight}}$ & {\color{OliveGreen}\textbf{+1.16}} & {\color{OliveGreen}\textbf{+1.28}} \\
        \midrule
        \rowcolor{LightGray} PARSeq [11] & 87.76 & 85.65 \\
        % \rowcolor{LightGray} PARSeq $d=480$ & +13.3 & +1.758 & +0.2 & {\color{OliveGreen}+0.1} & {\color{OliveGreen}+0.04} \\
        + $\text{CLIPTER}_{\text{Lightweight}}$ & {\color{OliveGreen}\textbf{+0.58}} & {\color{OliveGreen}\textbf{+0.69}} \\
        \bottomrule
    \end{tabular}
    }
    \vspace{-0.2cm}
\caption{
\textbf{CLIPTER Lightweight.} Even in its lightweight version, consisting only of a $\text{CLIP}_{\text{base}}$ image encoder and gated attention, CLIPTER enhances text recognizers.
% Results show that performance improvements originate from image-level context rather than model size.
}
\vspace{-0.7cm}
\label{tab:lightweight}
\end{center}
\end{table}

%% file: Tables/image_encoders.tex
\begin{table}
% \normalsize
\begin{center}
\bgroup
\small
        \resizebox{1\linewidth}{!}{%
        \begin{tabular}{lllcccc}
        \toprule
        & & \textbf{Image} & \multirow{2}{*}{\textbf{\textit{k}}} &  \textbf{Feature} &  \textbf{Average} & \textbf{Weighted} \\
        & & \textbf{Encoder} & & \textbf{Shape}  & 220,053 & \textbf{Average} \\
        \midrule
        \parbox[t]{1mm}{\multirow{8}{*}{\rotatebox[origin=c]{90}{\textbf{\small PARSeq}}}} & \mycc \parbox[t]{1mm}{\multirow{1}{*}{\rotatebox[origin=c]{90}{\textbf{\small Bl.}}}} & \mycc -- & \mycc -- & \mycc -- & \mycc 87.76 & \mycc 85.65 \\
        % \cmidrule{2-7}
        & \parbox[t]{2mm}{\multirow{3}{*}{\rotatebox[origin=c]{90}{\textbf{\small Vis.}}}} & DINO~\cite{caron2021emerging} & 2 & $ 50 \times 768$ & {\color{OliveGreen}+0.46} & {\color{OliveGreen}+0.58} \\
        & & ViT-MAE~\cite{he2022masked} & -- & $ 50 \times 1,024$ & {\color{OliveGreen}+0.44} & {\color{OliveGreen}+0.59} \\
        & & OWL-ViT~\cite{minderer2022simple} & 4 & $ 37 \times 768$ & {\color{OliveGreen}+0.55} & {\color{OliveGreen}+0.67} \\
        \cdashline{2-7}
        & \parbox[t]{2mm}{\multirow{4}{*}{\rotatebox[origin=c]{90}{\textbf{\small Vis.-Lan.}}}} & $\text{GIT}_{\text{L}}$~\cite{wang2022git} & 4 & $17 \times 1,024$ & {\color{OliveGreen}+0.53} & {\color{OliveGreen}+0.75} \\
        & & BLIP~\cite{li2022blip} & 5 & $ 37 \times 768$ & {\color{OliveGreen}+0.56} & {\color{OliveGreen}+0.74} \\
        & & $\text{CLIP}_{\text{base}}$~\cite{radford2021learning} & $\infty$ & $ 1 \times 512$ & {\color{OliveGreen}+0.69} & {\color{OliveGreen}+0.71} \\
        & & CLIP~\cite{radford2021learning} & $\infty$ & $1 \times 768$ & {\color{OliveGreen}\textbf{+0.73}} & {\color{OliveGreen}\textbf{+0.82}} \\
        % DONUT~\cite{kim2021donut} & $10\times10$ & 88.4 & 86.4 \\
        % DONUT~\cite{kim2021donut} box position & $10\times10$ & 88.3 & 86.3 \\
        \midrule
        \parbox[t]{1mm}{\multirow{5}{*}{\rotatebox[origin=c]{90}{\textbf{\small TRBA}}}} &
        \mycc \parbox[t]{1mm}{\multirow{1}{*}{\rotatebox[origin=c]{90}{\textbf{\small Bl.}}}} & \mycc -- & \mycc -- & \mycc -- & \mycc 86.28 & \mycc 83.38 \\
        % \cmidrule{2-7}
        & \parbox[t]{2mm}{\multirow{4}{*}{\rotatebox[origin=c]{90}{\textbf{\small Pooling}}}} 
        & BLIP~\cite{li2022blip} & $\infty$ & $ 1 \times 768$ & {\color{OliveGreen}+0.15} & {\color{OliveGreen}+0.12} \\
        & & BLIP~\cite{li2022blip} & 10 & $ 10 \times 768$ & {\color{OliveGreen}\textbf{+0.60}} & {\color{OliveGreen}\textbf{+0.87}} \\
        & & BLIP~\cite{li2022blip} & 5 & $ 37 \times 768$ & {\color{OliveGreen}+0.55} & {\color{OliveGreen}+0.80} \\
        & & BLIP~\cite{li2022blip} & 3 & $ 101 \times 768$ & {\color{OliveGreen}\textbf{+0.60}} & {\color{OliveGreen}+0.84} \\
        \bottomrule
    \end{tabular}
    }
\egroup
\vspace{-0.2cm}
% \tiny
\caption{\textbf{Image Encoder and Pooling.} Word accuracy for different pretrained image encoders, as well as pooling kernel sizes. \textbf{\small{Bl.}} stands for baseline }
\label{table:abla_image_encoders}
\vspace{-0.8cm}
\end{center}
\end{table}

%% file: Tables/integration_point.tex
\begin{table}
\begin{center}
\small
        \begin{tabular}{lcc}
        \toprule
        \multirow{2}{*}{\textbf{Method}} & \textbf{Average} & \textbf{Weighted} \\
        & 220,053 & \textbf{Average} \\
        \midrule
        \rowcolor{LightGray} TRBA \cite{Baek2019clova} & 86.28 & 83.38 \\
        % \cdashline{1-3}
        + $\text{\AlgoName}_{\text{Vision}}$ &  {\color{OliveGreen}+0.67} & {\color{OliveGreen}\textbf{+0.95}} \\
        + $\text{\AlgoName}_{\text{Contextual}}$ &  {\color{OliveGreen}+0.59} & {\color{OliveGreen}+0.9} \\
        + $\text{\AlgoName}_{\text{Decoder}}$ &  {\color{OliveGreen}\textbf{+0.72}} & {\color{OliveGreen}+0.9} \\
        \midrule
        \rowcolor{LightGray} ViTSTR-S \cite{vitstr_atienza2021vision} & 83.37 & 80.89 \\
        % \cdashline{1-3}
        + $\text{\AlgoName}_{\text{Vision}}$ & {\color{OliveGreen}\textbf{+1.2}} & {\color{OliveGreen}\textbf{+1.36}} \\
        \midrule
        \rowcolor{LightGray} ABINet-Vis \cite{abinet_fang2021read} & 82.14 & 79.75 \\
        % \cdashline{1-3}
        + $\text{\AlgoName}_{\text{Vision}}$ &  {\color{OliveGreen}\textbf{+1.73}} & {\color{OliveGreen}\textbf{+1.76}} \\
        + $\text{\AlgoName}_{\text{Contextual}}$ &  {\color{OliveGreen}+1.14} & {\color{OliveGreen}+0.97} \\
        \midrule
        \rowcolor{LightGray} ABINet \cite{abinet_fang2021read} & 85.85 & 83.85 \\
        % \cdashline{1-3}
        + $\text{\AlgoName}_{\text{Vision}}$ &  {\color{OliveGreen}+0.3} & {\color{OliveGreen}+0.18} \\
        + $\text{\AlgoName}_{\text{Contextual}}$  & {\color{OliveGreen}+0.18} & {\color{OliveGreen}+0.36} \\
        + $\text{\AlgoName}_{\text{Decoder}}$ & {\color{OliveGreen}\textbf{+0.49}} & {\color{OliveGreen}\textbf{+0.5}} \\
        \midrule
        \rowcolor{LightGray} PARSeq \cite{bautista2022parseq} & 87.76 & 85.65 \\
        % \cdashline{1-3}
        + $\text{\AlgoName}_{\text{Vision}}$ &  {\color{OliveGreen}{+0.55}} & {\color{OliveGreen}\textbf{+0.76}} \\
        + $\text{\AlgoName}_{\text{Decoder}}$ &  {\color{OliveGreen}\textbf{+0.56}} & {\color{OliveGreen}+0.71} \\
        \bottomrule
    \end{tabular}
    \vspace{-0.2cm}
\caption{
%\textbf{Word Accuracy when Varying Integration Point.} 
\textbf{Integration Point.}
In each text recognizer, there are several integration points to fuse the image and crop-level features. The results indicate that the optimal point depends on the recognizer architecture.}
\vspace{-0.9cm}
\label{tab:intergration_point}
\end{center}
\end{table}

%% file: Tables/ablations.tex
\begin{table}[t]
% \normalsize
\begin{center}
% \footnotesize
\bgroup
\small
        \resizebox{1\linewidth}{!}{%
        \begin{tabular}{lllccc}
        \toprule
        & \textbf{Fusion} & \textbf{Image} & \textbf{Recog.} & \textbf{Average} & \textbf{Weighted} \\
        & \textbf{Mechanism} & \textbf{Encoder} & \textbf{GFLOPS} & 220,053 & \textbf{Average} \\
        \midrule
        \parbox[t]{1mm}{\multirow{7}{*}{\rotatebox[origin=c]{90}{\textbf{\small TRBA}}}} & \mycc -- & \mycc -- & \mycc 6.681 & \mycc 86.28 & \mycc 83.38 \\
        % \cmidrule{2-6}
        & Gated attention & CLIP & +0.005 & {\color{OliveGreen}+0.62} & {\color{OliveGreen}+0.82} \\
        & MH-CA tiny & CLIP & +0.019 & {\color{OliveGreen}+0.50} & {\color{OliveGreen}+0.82} \\
        & MH-CA tiny & $\text{BLIP}_{k=5}$ & +0.033 & {\color{OliveGreen}+0.54} & {\color{OliveGreen}+0.77} \\
        & MH-CA mini & CLIP & +0.126 & {\color{OliveGreen}\textbf{+0.67}} & {\color{OliveGreen}\textbf{+0.95}} \\
        & MH-CA mini & $\text{BLIP}_{k=5}$ & +0.185 & {\color{OliveGreen}+0.56} &  {\color{OliveGreen}+0.86} \\
        & MH-CA small & CLIP & +0.502 & {\color{OliveGreen}+0.54} & {\color{OliveGreen}+0.81} \\
        & MH-CA small & $\text{BLIP}_{k=5}$ & +0.620 & {\color{OliveGreen}+0.52} & {\color{OliveGreen}+0.81} \\
        \midrule
        \parbox[t]{1mm}{\multirow{8}{*}{\rotatebox[origin=c]{90}{\textbf{\small ViTSTR}}}}
        & \mycc -- & \mycc -- & \mycc 4.608 & \mycc 83.37 & \mycc 80.89 \\
        % \cmidrule{2-6}
        & Gated attention & CLIP & +0.002 & {\color{OliveGreen}+1.22} & {\color{OliveGreen}+1.35} \\
        & MH-CA tiny  & CLIP & +0.004 & {\color{OliveGreen}\textbf{+1.25}} & {\color{OliveGreen}+1.34} \\
        & MH-CA tiny  & $\text{BLIP}_{k=5}$ & +0.016 & {\color{OliveGreen}+1.16} & {\color{OliveGreen}+1.29} \\
        & MH-CA mini  & CLIP & +0.024 & {\color{OliveGreen}+1.2} & {\color{OliveGreen}\textbf{+1.36}} \\
        & MH-CA mini  & $\text{BLIP}_{k=5}$ & +0.086 & {\color{OliveGreen}+1.21} & {\color{OliveGreen}+1.29} \\
        & MH-CA small & CLIP & +0.109 & {\color{OliveGreen}+1.16} & {\color{OliveGreen}+1.28} \\
        & MH-CA small & $\text{BLIP}_{k=5}$ & +0.223 & {\color{OliveGreen}+1.11} & {\color{OliveGreen}+1.26} \\
        \midrule
        \parbox[t]{1mm}{\multirow{8}{*}{\rotatebox[origin=c]{90}{\textbf{\small PARSeq}}}}
        & \mycc -- & \mycc -- & \mycc 3.174 & \mycc 87.76 & \mycc 85.65 \\
        % \cmidrule{2-6}
        & Gated attention & CLIP & +0.039 & {\color{OliveGreen}+0.53} & {\color{OliveGreen}+0.71} \\
        & MH-CA tiny  & CLIP & +0.081 & {\color{OliveGreen}+0.54} & {\color{OliveGreen}+0.71} \\
        & MH-CA tiny  & $\text{BLIP}_{k=5}$ & +0.098 & {\color{OliveGreen}+0.5} & {\color{OliveGreen}+0.7} \\
        & MH-CA mini  & CLIP & +0.533 & {\color{OliveGreen}+0.55} & {\color{OliveGreen}+0.76} \\
        & MH-CA mini  & $\text{BLIP}_{k=5}$ & +0.599 & {\color{OliveGreen}+0.56} & {\color{OliveGreen}+0.74} \\
        & MH-CA small & CLIP & +2.005 & {\color{OliveGreen}\textbf{+0.63}} & {\color{OliveGreen}\textbf{+0.77}} \\
        & MH-CA small & $\text{BLIP}_{k=5}$ & +2.137 & {\color{OliveGreen}+0.54} & {\color{OliveGreen}+0.72} \\
        \bottomrule
    \end{tabular}
    }
\egroup
\vspace{-0.2cm}
\caption{\textbf{Effect of the Fusion Mechanism.} Word accuracy when using Gated Attention or Multi-Headed Cross-Attention (MH-CA) fusion mechanisms. 
Note that the GFLOPS count refers to the recognition operation only.
% We found MH-CA mini as a balance between performance and computational load. 
%MH-CA stands for Multi-Headed Cross-Attention. 
}
\vspace{-0.7cm}
\label{table:fusion}
\end{center}
\end{table}

%% file: appendix_content.tex
\makeatletter
\newcommand\footnoteref[1]{\protected@xdef\@thefnmark{\ref{#1}}\@footnotemark}
\makeatother

\section{Pseudocode Algorithm}
\label{app:pseudocode}

The pseudocode for integrating CLIPTER into a recognizer is presented in \cref{alg:method}. This algorithm outlines the key components of our method, including image encoding, pooling, fusion mechanism, and integration point that divides the recognizer into encoder and decoder. In particular, the algorithm highlights that the image encoding operation is performed only once per image, regardless of its word count, and can be executed in parallel with the detection operation.

\input{pseudocode}

\section{Datasets}
\label{app:datasets}

\begin{figure*}[htp!]
    \centering
    \includegraphics[width=0.75\textwidth]%{Figures/datasets_examples.pdf}
    {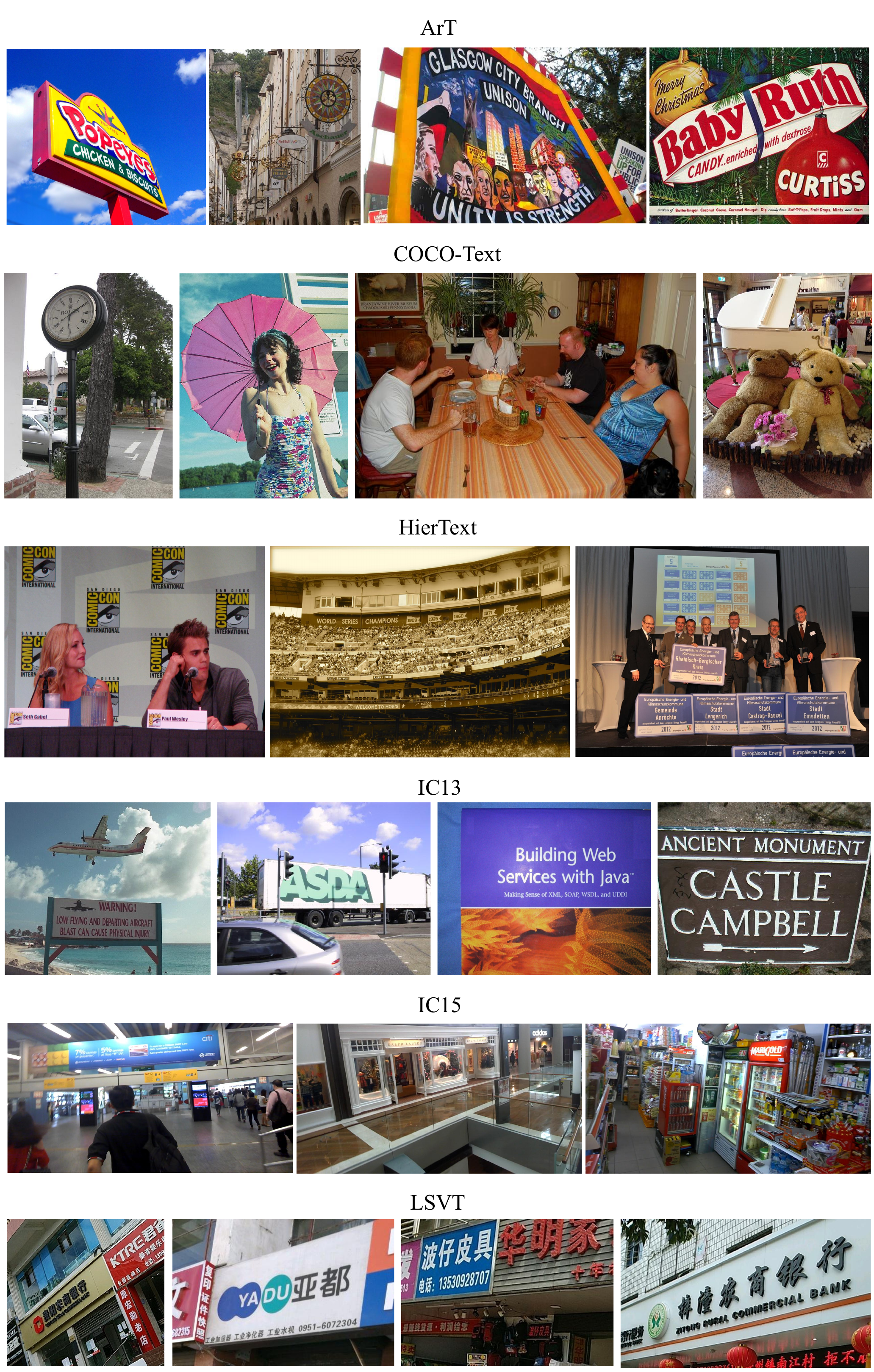}
    \caption{\textbf{Datasets Part 1.} We provide examples from each of the datasets used in this work.}
    \label{fig:dataset_sample_app_1}
    % \vspace{-0.5cm}
\end{figure*}

\begin{figure*}[htp!]
    \centering
    \includegraphics[width=0.77\textwidth]%{Figures/datasets_examples.pdf}
    {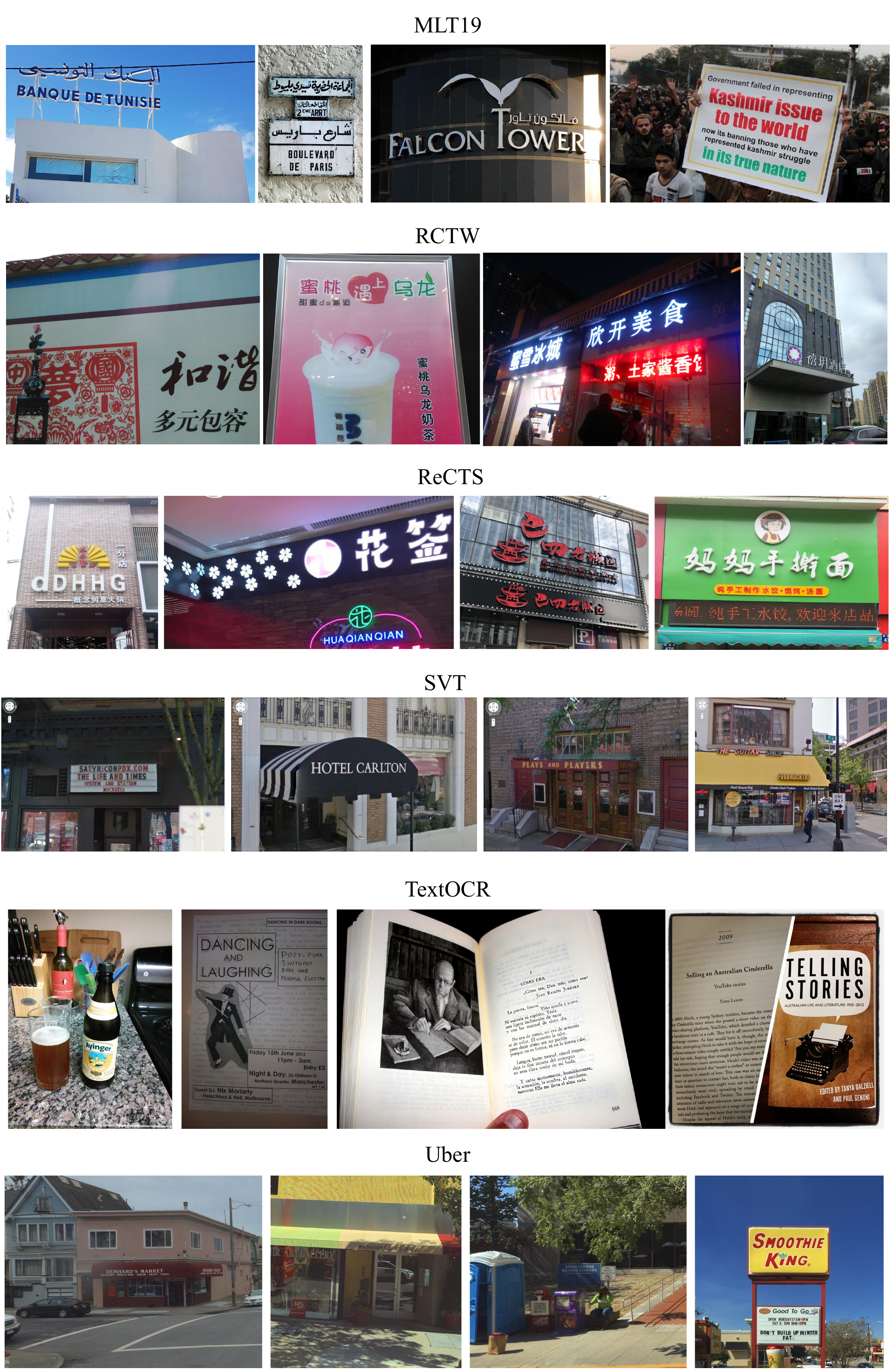}
    \caption{\textbf{Datasets Part 2.} We provide examples from each of the datasets used in this work.}
    \label{fig:dataset_sample_app_2}
    % \vspace{-0.5cm}
\end{figure*}

Our work utilizes a highly-diverse collection of 13 public benchmarks, depicted in \cref{fig:dataset_sample_app_1} and \cref{fig:dataset_sample_app_2}.
Since CLIPTER relies on the whole image together with the cropped words, we use datasets that have recognition and detection annotations, usually intended for the task of end-to-end text spotting. Therefore, we could not utilize some public test sets which contain only full images without localization annotations or cropped words without the full images. To mitigate this, we evaluate our method in these cases on the validation set or part of the training set. Nevertheless, we needed to omit IIIT-5k~\cite{Mishra2012sj} which contains only cropped text images and CUTE-80~\cite{Risnumawan2014cute} which does not contain end-to-end annotations. Below, we describe our data pre-processing and then, provide details on each dataset. 

\subsection{Data Pre-Processing}

Our work applies the same data filters on all datasets. In particular, we filter out words with the flag of \texttt{illegible} and words that have ignore labels, i.e., ``\#'', ``\#\#'', ``\#\#\#'', ``\#\#\#\#'' in general, ``.'' in TextOCR, and ``*'' in Uber. From the training data, we follow~ \cite{whatif_baek2021if} and also exclude text that consists of non-alphanumeric characters, long words that contain more than 25 characters, and vertical text by filtering words with more than two characters that their image height is greater than their image width.

\begin{figure*}[t!]
    \centering
    \begin{subfigure}[t]{0.5\textwidth}
        \centering
        \includegraphics[width=1.0\linewidth]{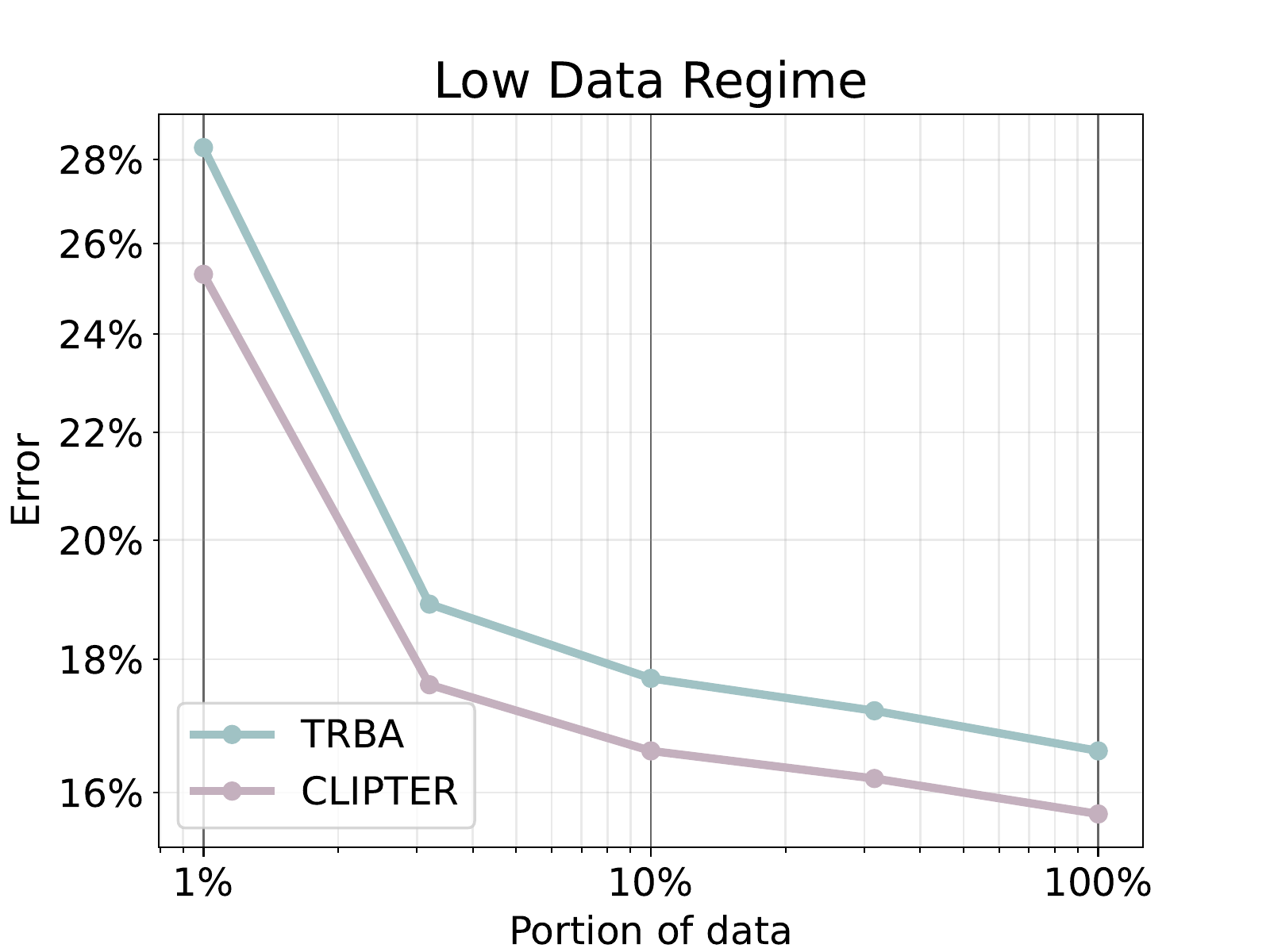}
    \end{subfigure}%
    ~ 
    \begin{subfigure}[t]{0.5\textwidth}
        \centering
        \includegraphics[width=1.0\linewidth]{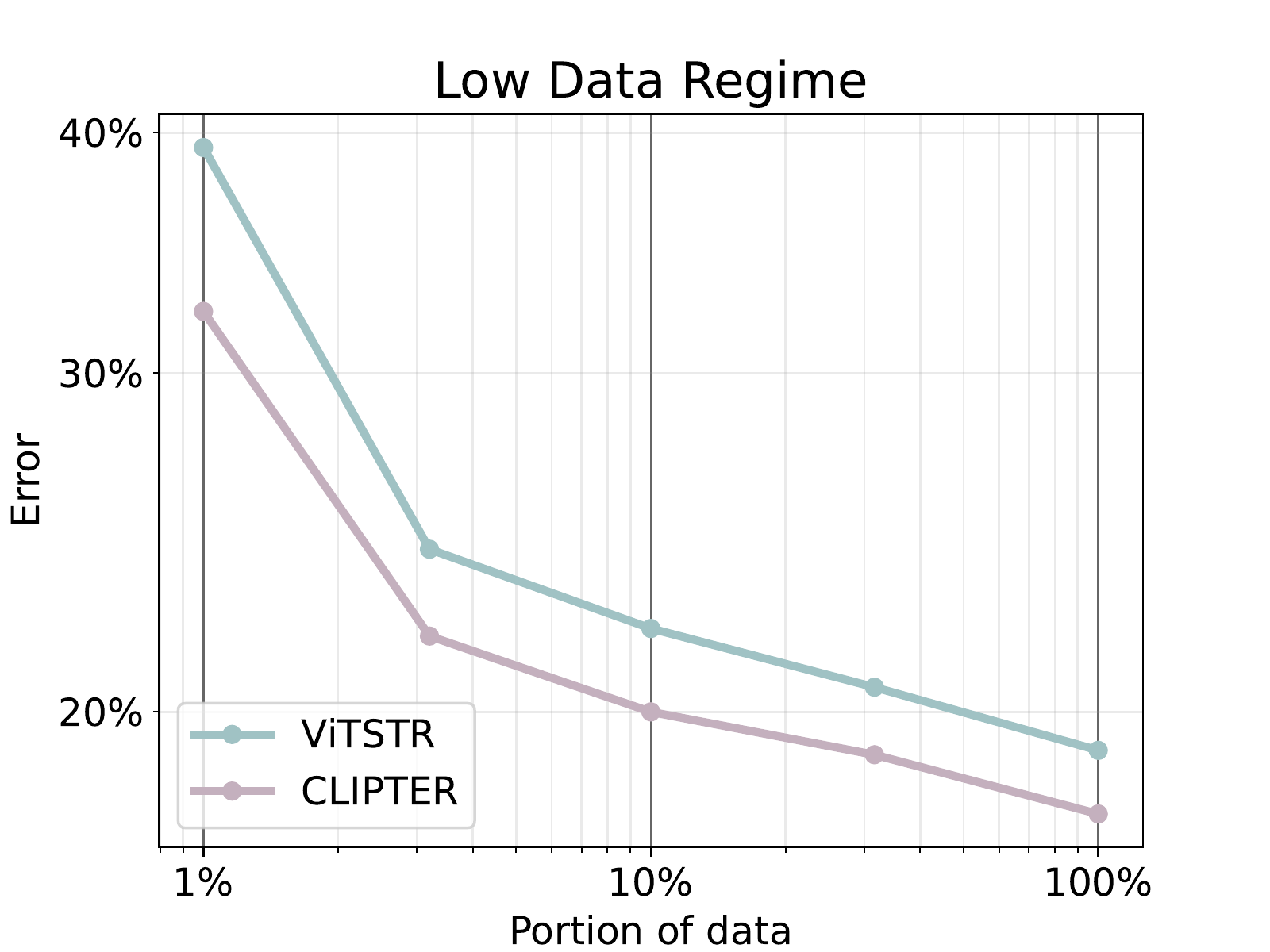}
    \end{subfigure}
    \caption{\textbf{Low Data Regime -- TRBA \& ViT-STR.} We evaluate the effect of CLIPTER with limited training data on TRBA~\cite{Baek2019clova} (left) and ViTSTR~\cite{vitstr_atienza2021vision} (right). Roughly speaking, adding CLIPTER to these architectures has more impact than doubling the training data amount in terms of reducing the error rate.}
    \label{fig:low_data_trba_vitstr}
\end{figure*}

\subsection{Dataset Details}

Below, we provide general details on each dataset and describe our data split into train, validation, and evaluation sets. A summary of these splits appears in \cref{table:datasets}, containing also data sizes. As we work on entire images as well as crops, we perform the splits at the entire image level.

\noindent\textbf{ArT}\cite{chng2019icdar2019} is a dataset of arbitrary-shaped text, collected from the train set of Task 3\footnote{\label{note1}\url{https://rrc.cvc.uab.es}}. The train set is divided into 80\% for training, 10\% for validation, and 10\% for evaluation.

\noindent\textbf{COCO-Text}\cite{veit2016coco} is based on COCO dataset~\footnote{\url{https://cocodataset.org}}, containing text in natural images\footnote{\url{https://vision.cornell.edu/se3/coco-text-2}}. We consider the training and validation sets that are published with bounding boxes, and split the training set into 90\% for training and 10\% for evaluation.

\noindent\textbf{HierText}\cite{long2022towards} features hierarchical annotations of text in natural scenes and documents\footnote{\url{https://github.com/google-research-datasets/hiertext}}. We consider the training and validation sets which have available bounding boxes, and split the training set into 90\% for training and 10\% for evaluation. In this dataset, we filtered words that are annotated as vertical.

\noindent\textbf{IC13}\cite{Karatzas2013ic13} contains images that are focused around the text content\footnoteref{note1}. Since only the training set is provided with full annotations, we use it all for evaluation.

\noindent\textbf{IC15}\cite{Karatzas2015ic15} contains incidental scene text and therefore is more challenging\footnoteref{note1}. The test set here is the official one, while the training set is divided into 90\% for training and 10\% for validation.

\noindent\textbf{LSVT}\cite{lsvt_sun2019chinese} contains scene text in street view images\footnoteref{note1}. Here, only the training set has full annotations. Therefore, we divide it into 80\% for training 10\% for validation, and 10\% for evaluation.

\noindent\textbf{MLT19}\cite{nayef2019icdar2019} is a multilingual dataset\footnoteref{note1}. The training set is divided into language subsets, from which we consider English, French, German, and Italian. We split these data into 80\% for training, 10\% for validation, and 10\% for evaluation.

\input{Tables/datasets}

\noindent\textbf{OOV}\cite{garcia2022out} is a new dataset containing out-of-vocabulary scene text\footnoteref{note1}. Since this dataset is based on other datasets, we did not use its training set, but use its validation set for evaluation. In this dataset, we filter words that are annotated as non-English or vertical.

\noindent\textbf{RCTW}\cite{shi2017icdar2017} is a dataset for reading Chinese text in images\footnote{\url{https://rctw.vlrlab.net}}. We split the published training set in 80\% for training, 10\% for validation and 10\% for evaluation.

\noindent\textbf{ReCTS}\cite{zhang2019icdar} contains Chinese text on signboard\footnoteref{note1}. We split the published training set in 80\% for training, 10\% for validation and 10\% for evaluation. In this dataset, we ignore words that are annotated with the flag of \texttt{ignore}.

\noindent\textbf{SVT}\cite{wang2011end} contains street view text in images from Google Street View\footnote{\url{https://tc11.cvc.uab.es/datasets/SVT\_1}}. Here, we use the official test set and divide the training set into 90\% for training and 10\% for validation.

\noindent\textbf{TextOCR}\cite{singh2021textocr} contains high quality images from OpenImages\footnote{\url{https://storage.googleapis.com/openimages/web/index.html}} with an average of 30 words per image\footnote{\url{https://textvqa.org/textocr}}. Here, we use the published validation set and divide the training set into 90\% for training and 10\% for evaluation.

\noindent\textbf{Uber}\cite{zhang2017uber} contains street-level images collected from car mounted sensors\footnote{\url{https://s3-us-west-2.amazonaws.com/uber-common-public/ubertext/index.html}}. We keep the original split of training, validation, and evaluation sets.

\begin{figure*}[t]
    \centering
    % Subfigure 1
    \begin{subfigure}{0.138\linewidth}
        \centering
        \includegraphics[width=\linewidth]{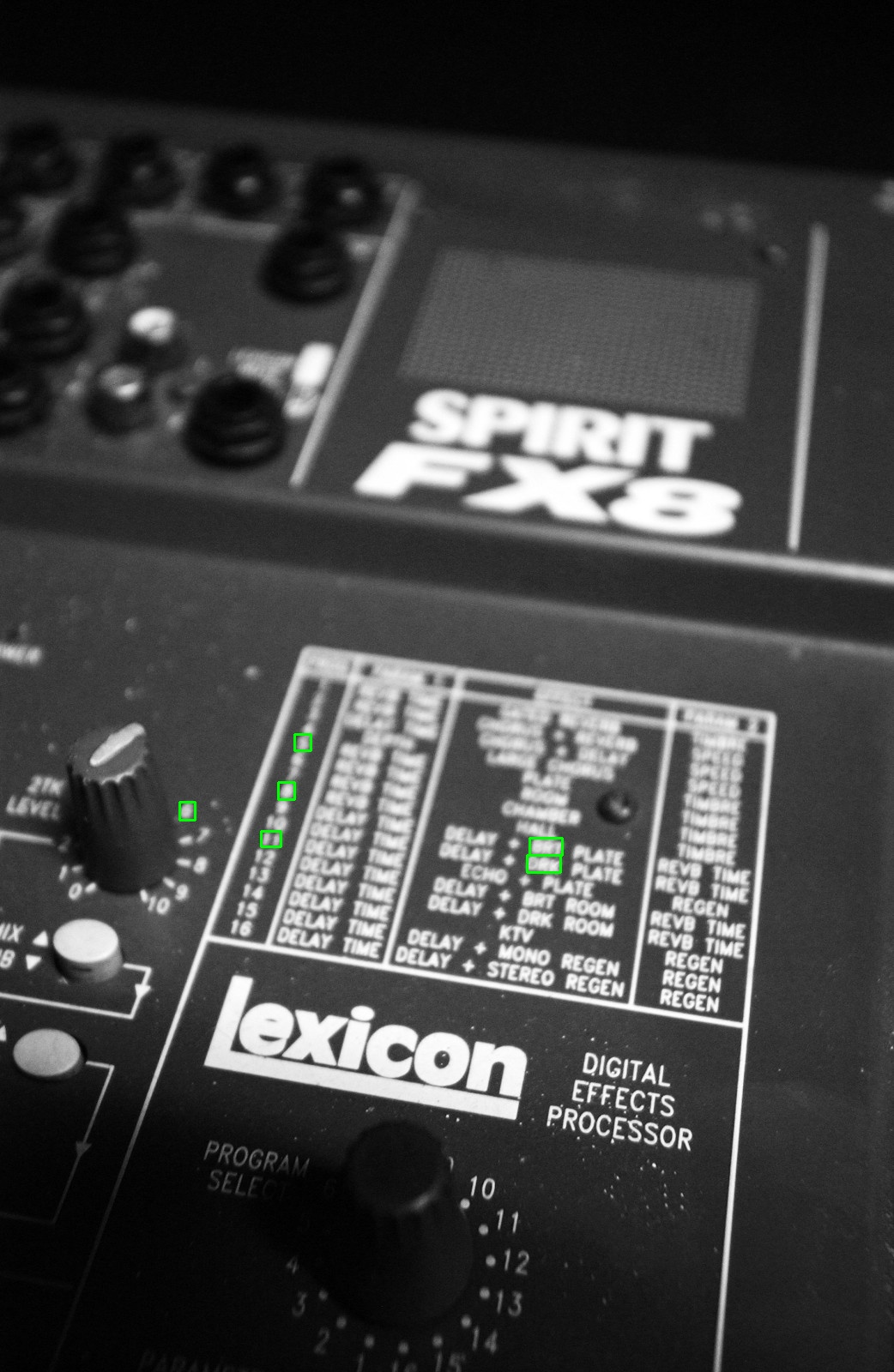}
    \end{subfigure}
    % Subfigure 4
    \begin{subfigure}{0.212\linewidth}
        \centering
        \includegraphics[width=\linewidth]{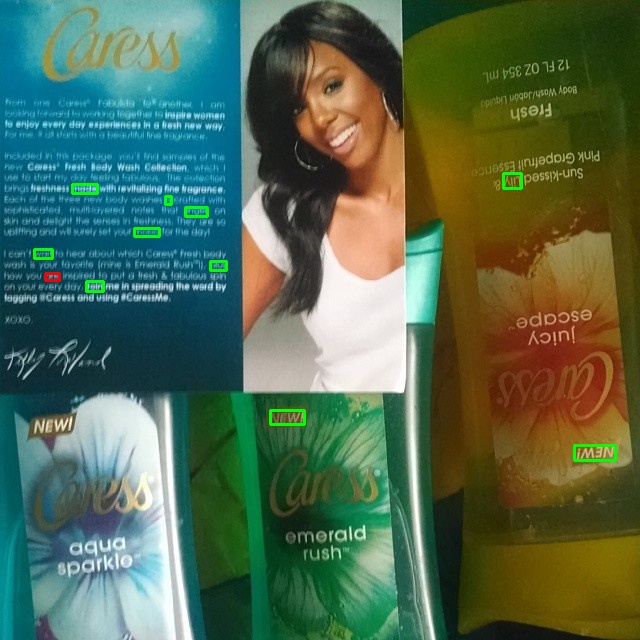}
    \end{subfigure}
    % Subfigure 2
    \begin{subfigure}{0.281\linewidth}
        \centering
        \includegraphics[width=\linewidth]{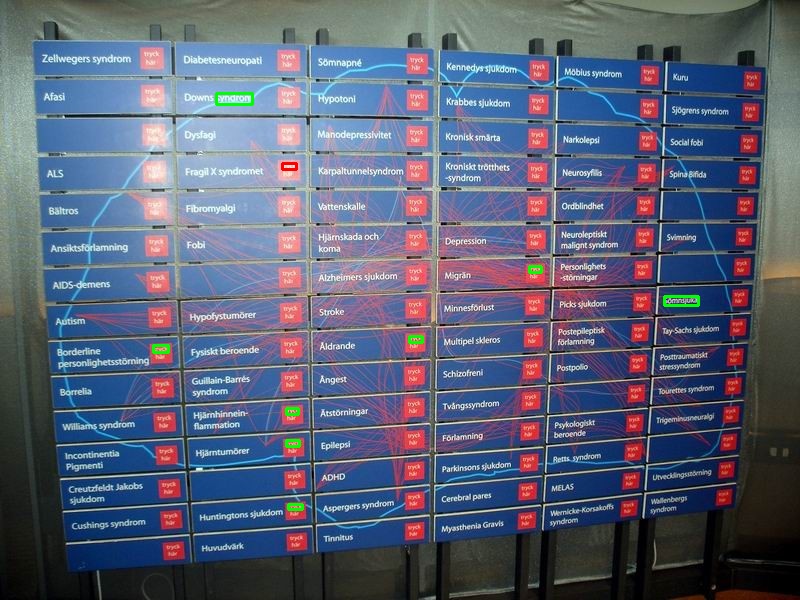}
    \end{subfigure}
    % Subfigure 3
    \begin{subfigure}{0.318\linewidth}
        \centering
        \includegraphics[width=\linewidth]{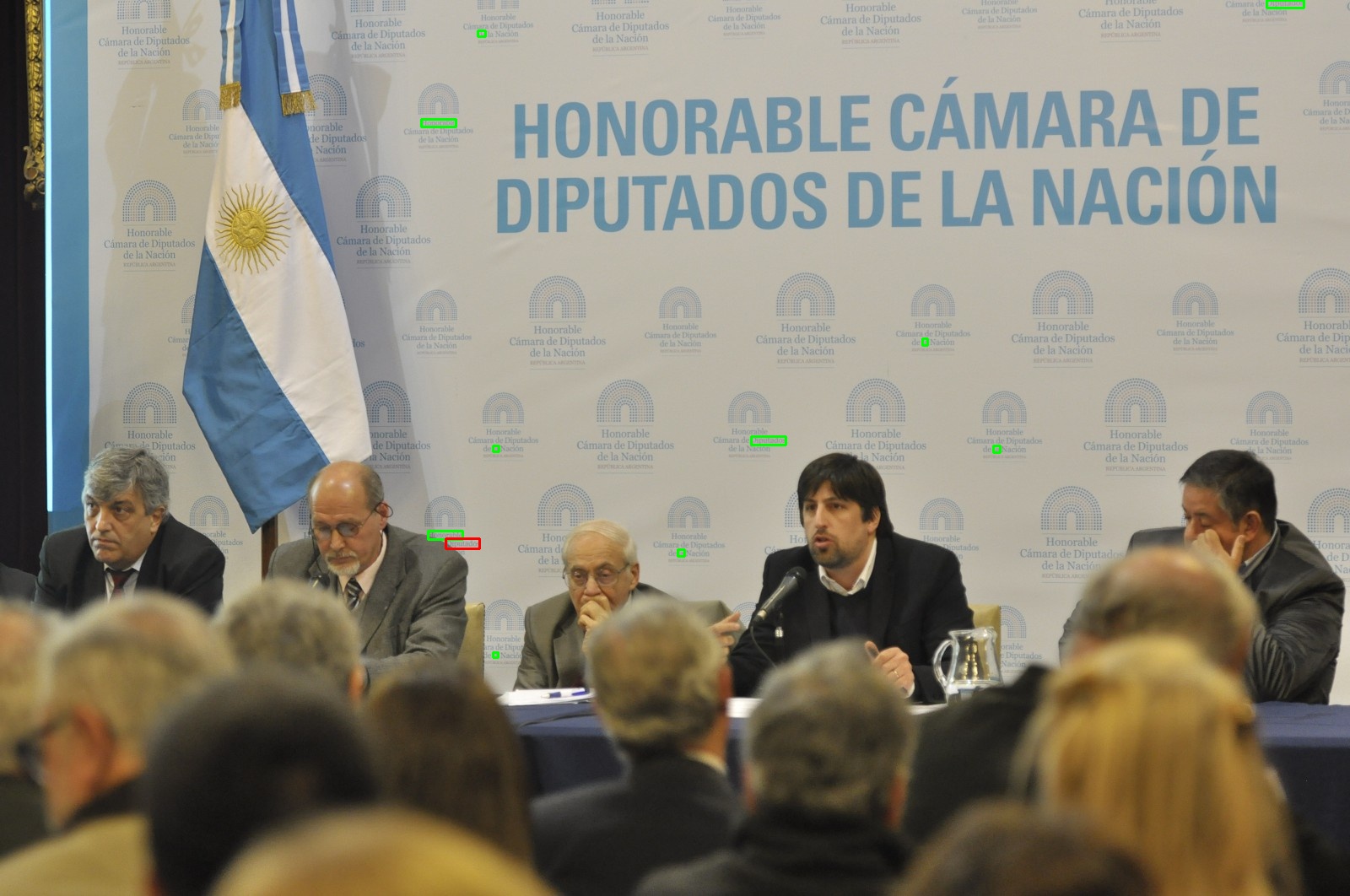}
    \end{subfigure}
    \caption{\textbf{Quantitative Results on Rich-in-Text Images.} Images with dense text (\textgreater100) that benefit from integrating scene-level information using CLIPTER. Green boxes highlight words accurately transcribed by PARSeq+CLIPTER but not by PARSeq, while red boxes indicate the opposite.}
    \label{fig:examples}
\end{figure*}

\section{Implementation Details}
\label{app:implementation_details}

\paragraph{Multi-head Cross-Attention fusion mechanism.} Our implementation of the Multi-Head Cross-Attention (MH-CA) mechanism is based on the implementation of BERT \cite{devlin2018bert,bhargava2021generalization} proposed by HuggingFace. Table \ref{tab:ca_size} presents further architectural details.

\paragraph{Training details.} Baseline STR models are trained with the hyperparameters published by respective authors. CLIPTER is trained for 20 epochs with a learning rate varying from ${1\times10^{-5}}$ to ${3\times10^{-5}}$. Specifically, gated-attention, MH-CA tiny, mini and small are trained with learning rates of ${2\times10^{-5}}$, ${3\times10^{-5}}$, ${3\times10^{-5}}$ and ${1\times10^{-5}}$ respectively.

\input{Tables/ca_size}
\input{Tables/Synthetic_data} 
\input{Tables/uber_category}

\section{Low Data Regime}
\label{app:low_data}

Similarly to analysis performed in the main paper over PARSeq, we evaluate the effect of our method in the low data regimes on TRBA and ViTSTR architectures. As shown in \cref{fig:low_data_trba_vitstr}, utilizing CLIPTER on these schemes achieves better results than the baseline model with doubled amount of training data.

\section{Latency Analysis}
\label{app:latency}

To evaluate the impact of our solution on recognition latency, we conduct end-to-end (E2E) experiments on the ICDAR-15 and Total-Text datasets, and calculate the frames per second (FPS).
To this end, we use the ResNet50-based detection model from GLASS~\cite{ronen2022glass}\footnote{\url{https://github.com/amazon-science/glass-text-spotting}} and exclude their recognition components.
Our experiments are conducted on a single V100 NVidia GPU and a simple PyTorch implementation, without any optimizations, such as TensorRT, that could improve the latency results.
We calculate the latency using PyTorch benchmarking code\footnote{\url{https://pytorch.org/tutorials/recipes/recipes/benchmark.html\#pytorch-benchmark}}, with FPS calculated as the average of the median run-time per image.
Evaluation metrics are in accordance with the protocol of \cite{ronen2022glass}.

\section{Additional Experiments} 

\subsection{Synthetic Data} 
In this part, we aim to analyze the effect of utilizing synthetic data. To this end, we train PARSeq with and without CLIPTER also on the large synthetic datasets of MJ~\cite{jaderberg2014synthetic} and ST~\cite{gupta2016synthetic}. As shown in \cref{table:synthetic}, adding the large synthetic data, about 14M images, to the training set only marginally improves the results, indicating on the low impact of synthetic data when there is a lot of real-world data. That said, these datasets do lead to significant improvements on IC13 and IC15. This finding, revealed also in~\cite{aberdam2022multimodal}, indicates that these datasets mainly represent specific types of natural scenarios.

\begin{figure}[t]
    \centering
    \includegraphics[width=1.0\linewidth]{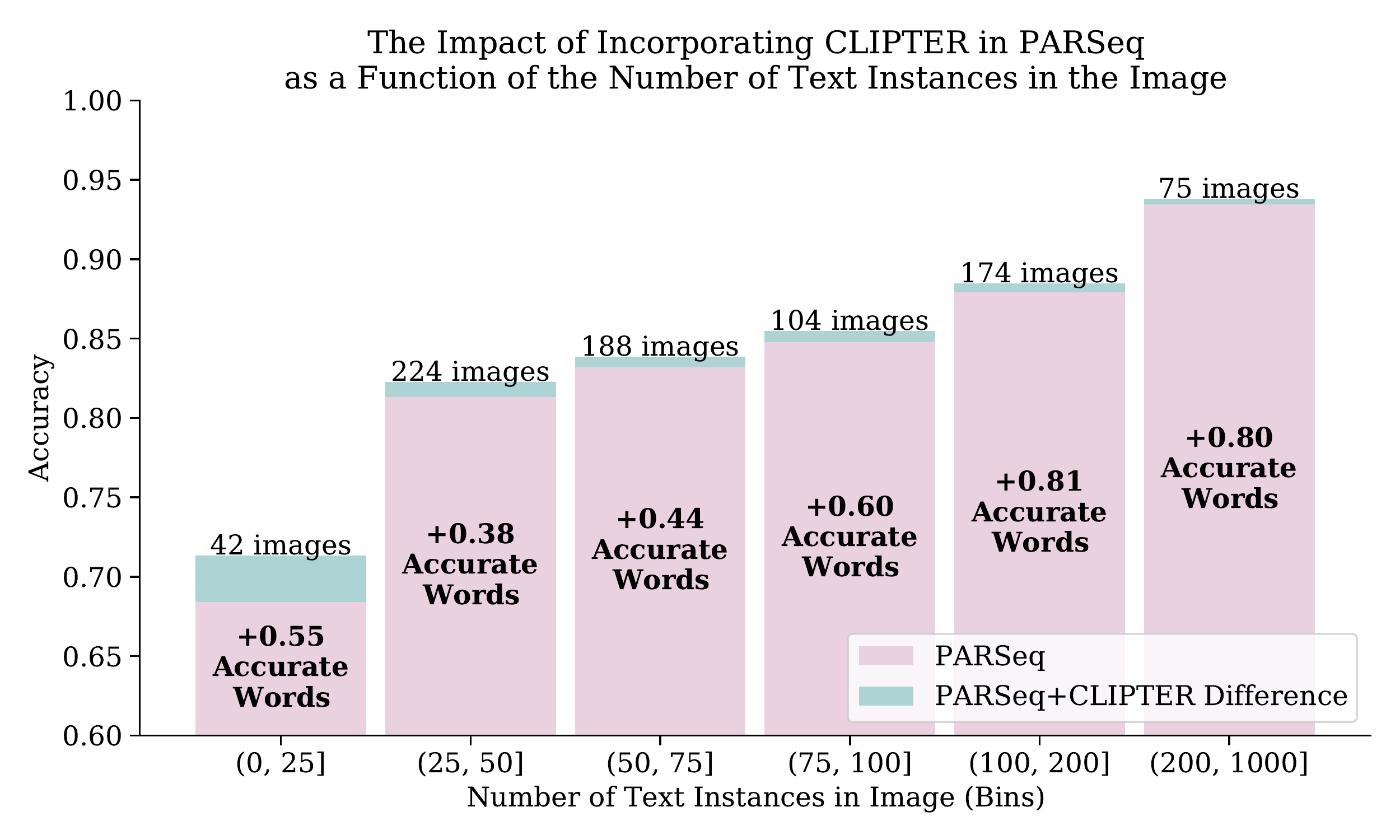}
    \vspace{-0.6cm}
    \caption{\textbf{Enhancing Performance in Dense-Text Images.} This figure illustrates the averaged improvement in accuracy and the number of accurately transcribed words relative to the total number of words in the image. Our algorithm demonstrates remarkable success even in densely-packed text images.}
    \label{fig:plot}
\end{figure}

\subsection{Breaking-Down Results on Uber-Text} 

We utilize Uber-Text~\cite{zhang2017uber} word categories to break down the results of PARSeq with and without CLIPTER. As shown in \cref{tab:uber}, our method is especially efficient on business name (+1.3\%) and street numbers (+1.3\%). We believe that these improvements are thanks to the use of a vision-language model that was pretrained also on the textual descriptions of the images, which often contain such information as it is crucial for understanding the scene.

\subsection{Dense Documents}
\label{app:dense_documents}

We conduct both a quantitative (Figure \ref{fig:plot}) and qualitative (Figure \ref{fig:examples}) analysis on the text-dense HierText dataset.
The results demonstrate that our model consistently improves accuracy, even in highly text-dense images with over 100 words.

\section{Further qualitative analysis}
\label{app:qualitative}
\begin{figure*}[htp!]
    \centering
    \includegraphics[width=0.8\textwidth]{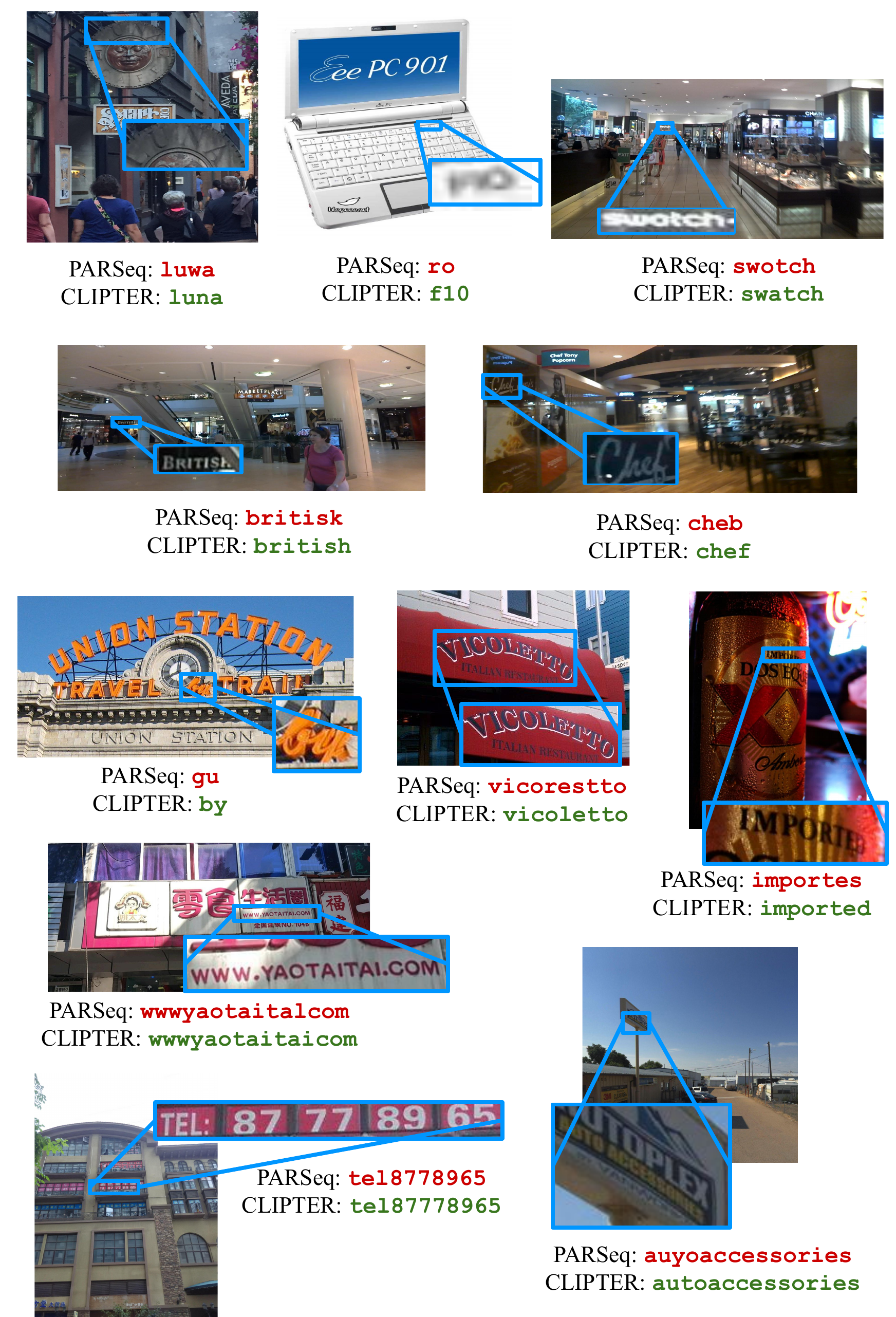}
    \caption{\textbf{Positive flips.} Examples in which CLIPTER corrected the prediction of PARSeq and matched the GT annotation.}
    \label{fig:positive_flips}
    % \vspace{-0.5cm}
\end{figure*}

\cref{fig:positive_flips} displays additional examples showcasing benefits of CLIPTER.

\label{fig:pos_flips}

\begin{figure*}[htp!]
    \centering
    \includegraphics[width=0.8\textwidth]{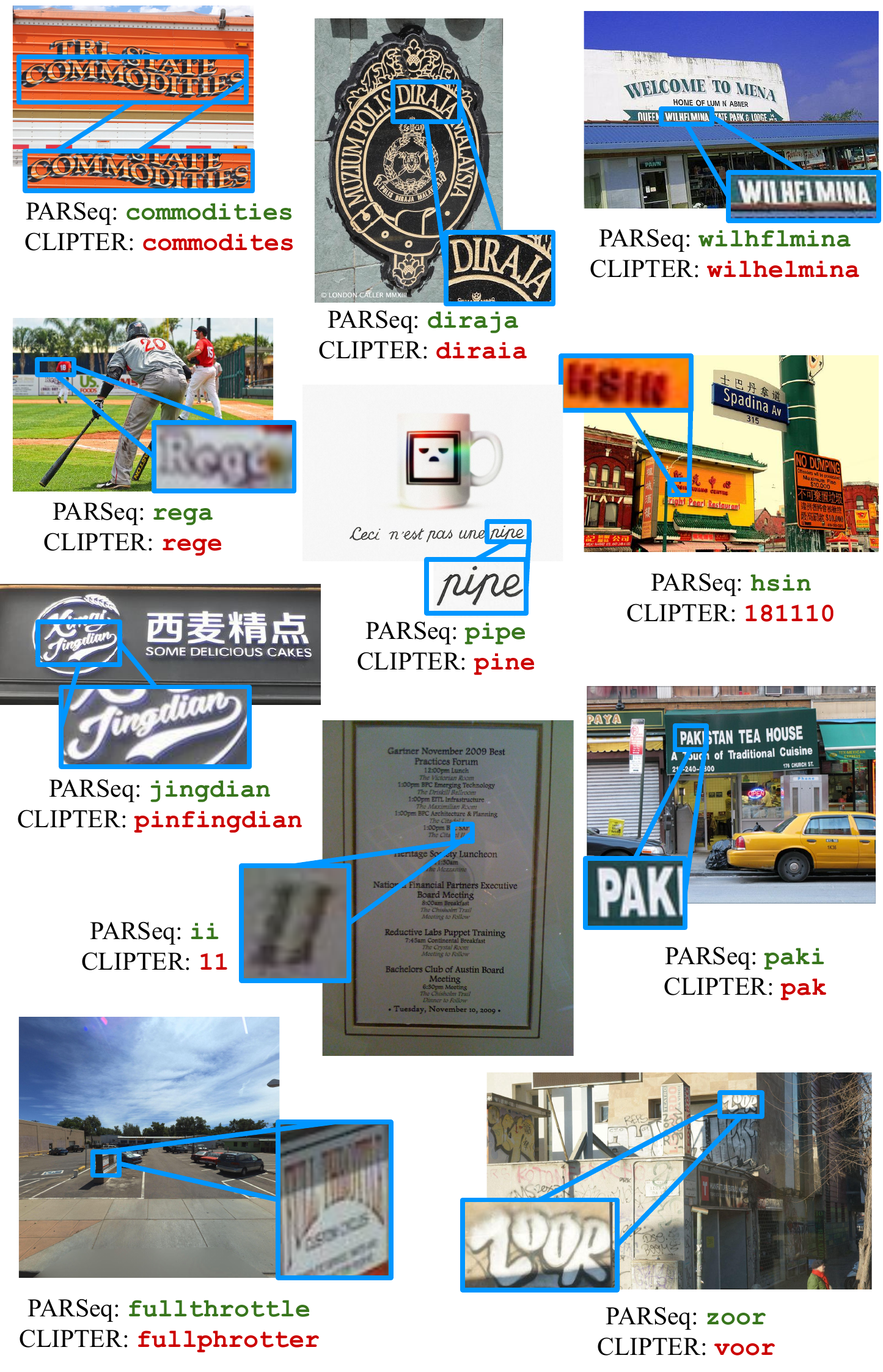}
    \caption{\textbf{Negative flips.} Examples in which CLIPTER harmed the prediction of PARSeq which previously matched the GT annotation.}
    \label{fig:negative_flips}
    % \vspace{-0.5cm}
\end{figure*}
\label{fig:neg_flips}

\clearpage

%% file: pseudocode.tex
\begin{center}
\begin{algorithm}[h]
  \caption{CLIPTER PyTorch-like pseudocode}
  \label{alg:method}
    \definecolor{codeblue}{rgb}{0.25,0.5,0.5}
    \definecolor{codekw}{rgb}{0.85, 0.18, 0.50}
    \newcommand{\algofontsize}{8.0pt}
    \lstset{
      backgroundcolor=\color{white},
      basicstyle=\fontsize{\algofontsize}{\algofontsize}\ttfamily\selectfont,
      columns=fullflexible,
      breaklines=true,
      captionpos=b,
      commentstyle=\fontsize{\algofontsize}{\algofontsize}\color{codeblue},
      keywordstyle=\fontsize{\algofontsize}{\algofontsize}\color{black},
      xleftmargin=-0.25cm,
    }
    
\begin{lstlisting}[language=python]
"""
img: scene image
text_crops: all text images cropped from image
img_encoder: frozen VL image encoder
k: kernel of average pooling
fusion_ca: nn.MultiHeadAttention()
alpha: gated parameter (init as 0)
recog_encdoer, recog_decoder: the recognition modules before and after the integation point
"""

# image encoding (in parallel to detection)
with torch.no_grad():
    img_f = img_encoder(img)  # (1 + HW, d)
    img_f = [img_f[0], avg_pool2d(img_f[1:], k)]

preds = []
for crop in text_crops:
    # recognizer encoding
    crop_f = recog_encoder(crop)

    # fusion by gated cross attention
    merged_f = fusion_ca(query=crop_f, key=img_f, value=img_f)
    c = torch.tanh(alpha)
    fused_f = (1 - c) * crop_f + c * merged_f

    # recognizer decoding
    preds.append(recog_decoder(fused_f))
\end{lstlisting}
\end{algorithm}
\vspace{-1cm}
\end{center}

%% file: Tables/datasets.tex
\begin{table}[t]
\normalsize
\begin{center}
\footnotesize
\bgroup
\def\arraystretch{1.1}
        % \resizebox{1\textwidth}{!}{%
    %     \begin{tabular}{lcccccccccccc|c}
    %     \toprule
    %     \textbf{Datasets} & \textbf{SVT} & \textbf{IC13} & \textbf{IC15} & \textbf{COCO-Text} & \textbf{RCTW} & \textbf{Uber} & \textbf{ArT} & \textbf{LSVT} & \textbf{ReCTS} & \textbf{MLT19} & \textbf{TextOCR} & \textbf{HierText} & \textbf{All}
    %     \\
    %     \midrule
    %     \#Raw word boxes & 914 & 848 & 6545 & 146K & 65K & 285K & 50K & 384K & 147K & 89K & 903K & 1,200K & 3,277K \\
    %     \midrule
    %     \#Processed word boxes & 903 & 757 & 6167 & 70K & 9816 & 155K & 31K & 40K & 23K & 42K & 733K & 940K & 2,052K \\ \midrule \midrule
    %     Train size & 232 & 0 & 3741& 51K & 7,837& 75K & 25K & 32K & 18K & 34K & 566K & 711K & 1,516K \\
    %     \midrule
    %     Validation size & 24 & 0 & 349 & 13K & 1017 & 30K & 2,701 & 3,937 & 2,331 & 3,970 & 96K & 163K & 316K \\
    %     \midrule
    %     Test size & 647 & 757 & 2,077 & 5,716 & 962 & 50K & 3,667 & 3,911 & 2,219 & 4,100 & 71K & 76K & 220K \\
    %     \bottomrule
    % \end{tabular}
    
    \begin{tabular}{lcccccc}
    \toprule
    & \multicolumn{3}{c}{Public E2E Annotations} & \multicolumn{3}{c}{Number of Words} \\
    & Train. & Valid. & Eval. & Train. & Valid. & Eval. \\
    \midrule
    {ArT} & \checkmark & \ding{55} & \ding{55} & 25K & 2,701 & 3,667 \\
    {COCO-Text} & \checkmark & \checkmark & \ding{55} & 51K & 13K & 5,716 \\
    {HierText} & \checkmark & \checkmark & \ding{55} & 711K & 163K & 76K \\
    {IC13} & \checkmark & \ding{55} & \ding{55} & -- & -- & 757 \\
    {IC15} & \checkmark & \ding{55} & \checkmark & 3,741 & 349 & 2,077 \\
    {LSVT} & \checkmark & \ding{55} & \ding{55} & 32K & 3,937 & 3,911 \\
    {MLT19} & \checkmark & \ding{55} & \ding{55} & 34K & 3,970 & 4,100 \\
    % {OOV} & \\
    {RCTW} & \checkmark & \ding{55} & \ding{55} & 7,837 & 1,017 & 962 \\
    {ReCTS} & \checkmark & \ding{55} & \ding{55} & 18K & 2,331 & 2,219 \\
    {SVT} & \checkmark & \ding{55} & \checkmark & 232 & 24 & 647 \\
    {TextOCR} & \checkmark & \checkmark & \ding{55} & 566K & 96K & 71K \\
    {Uber} & \checkmark & \checkmark & \checkmark & 75K & 30K & 50K \\
    \midrule
    \textbf{All} & & & & \textbf{1,516K} & \textbf{316K} & \textbf{220K} \\
    \bottomrule
    \end{tabular}
    % }
\egroup
% \vspace{-0.2cm}
% \tiny
\caption{\textbf{Dataset Partition.} Number of cropped word images after pre-processing and splitting into training, validation, and evaluation sets.}
\label{table:datasets}
% \vspace{-0.8cm}
\end{center}
\end{table}

%% file: Tables/ca_size.tex
\begin{table}[t]
    \centering
    \small
    \resizebox{1\linewidth}{!}{%
    \begin{tabular}{lccccc}
    \toprule
        \multirow{2}{*}{CA Model} & \# Attention & \# Hidden & Hidden & Intermediate & \multirow{2}{*}{\# Parameters} \\
        & Heads & Layers & Size & Size \\
        \midrule
        Gated-Attention & -- & -- & -- & -- & 328K\\
        MH-CA Tiny & 2 & 2 & 128 & 512 & 923K \\
        MH-CA Mini & 4 & 4 & 256 & 1,024 & 5.3M \\
        MH-CA Small & 8 & 4 & 512 & 2,048 & 18.1M \\
    \bottomrule
    \end{tabular}
    }
    \vspace{-0.2cm}
    \caption{\textbf{Cross-Attention Model Size.}}
    % \vspace{-0.3cm}
    \label{tab:ca_size}
\end{table}

%% file: Tables/Synthetic_data.tex
\begin{table*}
\normalsize
\begin{center}
\footnotesize
\bgroup
\def\arraystretch{1.1}
    %     \resizebox{1\textwidth}{!}{%
    %     \begin{tabular}{lccccccccccccc}
    %     \toprule
    %     \textbf{Datasets} & \textbf{SVT} \cite{wang2011end} & \textbf{IC13}\cite{Karatzas2013ic13} & \textbf{IC15}\cite{Karatzas2015ic15}& \textbf{COCO-Text} \cite{veit2016coco} & \textbf{RCTW} \cite{shi2017icdar2017}& \textbf{Uber} \cite{zhang2017uber} & \textbf{ArT} \cite{chng2019icdar2019} & \textbf{LSVT} \cite{Phan2013svtp} & \textbf{ReCTS} \cite{zhang2019icdar} & \textbf{MLT19}\cite{nayef2019icdar2019} & \textbf{TextOCR} \cite{singh2021textocr} & \textbf{HierText} \cite{long2022towards} & \textbf{All}
    %     \\
    %     \midrule
    %     Parseq - Real Data & 96.1 & 98.9 & 85.7 & 80.5 & 81.4 & 83.2 & 91.2 & 80.2 & 91.8 & 91.5 & 85.2 & 87.4 & 85.6 \\
    %     \midrule
    %     Parseq - Real Data + Synthetic Data & 97.2 & 99.5 & 86.4 & 80.6 & 83 & 82.1 & 91.1 & 80.2 & 91.9 & 91.7 & 85.1 & 87.5 & 85.5 \\ \midrule 
    %     \midrule
    %     CLIPTER - Real Data & 97.8 & 99.5 & 86.7 & 81.4 & 83.6 & 83.1 & 91.4 & 81.3 & 92.6 & 92 & 85.9 & 88.4 & 86.3 \\
    %     \midrule
    %     CLIPTER - Real Data + Synthetic Data & 96.6 & 99.1 & 85.9 & 81 & 82.1 & 84.4 & 91.7 & 82 & 91.8 & 91.6 & 86 & 88 & 86.4 \\
    %     \bottomrule
    % \end{tabular}
    % }
        \resizebox{1\textwidth}{!}{%
        \begin{tabular}{llcccccccccccccc}
        \toprule
% \begin{table*}[!t]
% 	\centering
% 	\resizebox{1\textwidth}{!}{%
% 	\begin{tabular}	{@{\extracolsep{1pt}}lcccccccccccccc}
%         \toprule	 	
        & \multirow{2}{*}{\textbf{Method}} &  \textbf{SVT} & \textbf{IC13} & \textbf{IC15} & \textbf{COCO} & \textbf{RCTW} & \textbf{Uber} & \textbf{ArT} & \textbf{LSVT} & \textbf{RECTS} & \textbf{MLT19} & \textbf{TextOCR} & \textbf{HierText} & \textbf{Average} & \textbf{Weighted} \\
        & & 647 & 757 & 2,077 & 5,716 & 962 & 49,561 & 3,677 & 3,911 & 2,219 & 4,100 & 70,597 & 75,829 & 220,053 & \textbf{Average} \\
        \midrule
        \parbox[t]{1mm}{\multirow{3}{*}{\rotatebox[origin=c]{90}{\textbf{\small Real}}}} & \mycc PARSeq \cite{bautista2022parseq} & \mycc 96.1 & \mycc 98.9 & \mycc 85.7 & \mycc 80.5 & \mycc 81.4 & \mycc 83.2 & \mycc 91.2 & \mycc 80.2 & \mycc 91.8 & \mycc 91.5 & \mycc 85.2 & \mycc 87.4 & \mycc 87.8 & \mycc 85.6 \\
        & + $\text{\AlgoName}_{\text{Vision}}$ & {96.6} & {99.1} & {85.9} & {81.0} & {82.1} & \textbf{84.4} & \textbf{91.7} & \textbf{81.8} & {91.8} & {91.6} & \textbf{86.0} & {88.0} & {88.3} & \textbf{86.4} \\
        & $\quad \quad \Delta$ & {\color{OliveGreen}\textbf{+0.5}} & {\color{OliveGreen}\textbf{+0.2}} & {\color{OliveGreen}\textbf{+0.2}} & {\color{OliveGreen}\textbf{+0.5}} & {\color{OliveGreen}\textbf{+0.7}} & {\color{OliveGreen}\textbf{+1.2}} & {\color{OliveGreen}\textbf{+0.5}} & {\color{OliveGreen}\textbf{+1.6}} & 
        0 &
        {\color{OliveGreen}\textbf{+0.1}} & {\color{OliveGreen}\textbf{+0.8}} & {\color{OliveGreen}\textbf{+0.6}} & {\color{OliveGreen}\textbf{+0.5}} & {\color{OliveGreen}\textbf{+0.8}} \\
        \midrule
        \parbox[t]{1mm}{\multirow{3}{*}{\rotatebox[origin=c]{90}{\textbf{\small + Synth.}}}} & \mycc PARSeq \cite{bautista2022parseq} & \mycc 97.2 & \mycc \textbf{99.5} & \mycc 86.4 & \mycc 80.6 & \mycc 82.8 & \mycc 82.1 & \mycc 91.1 & \mycc 80.2 & \mycc 91.9 & \mycc 91.7 & \mycc 85.1 & \mycc 87.5 & \mycc 88.0 & \mycc 85.4 \\
        & + $\text{\AlgoName}_{\text{Vision}}$ & \textbf{97.8} & \textbf{99.5} & \textbf{86.7} & \textbf{81.4} & \textbf{83.6} & 83.1 & 91.4 & 81.3 & \textbf{92.6} & \textbf{92.0} & 85.9 & \textbf{88.4} & \textbf{88.6} & 86.3 \\
        & $\quad \quad \Delta$ & {\color{OliveGreen}\textbf{+0.6}} & 0 & {\color{OliveGreen}\textbf{+0.3}} & {\color{OliveGreen}\textbf{+0.8}} & {\color{OliveGreen}\textbf{+0.8}} & {\color{OliveGreen}\textbf{+1.0}} & {\color{OliveGreen}\textbf{+0.3}} & {\color{OliveGreen}\textbf{+1.1}} & 
        {\color{OliveGreen}\textbf{+0.7}} &
        {\color{OliveGreen}\textbf{+0.3}} & {\color{OliveGreen}\textbf{+0.8}} & {\color{OliveGreen}\textbf{+0.9}} & {\color{OliveGreen}\textbf{+0.6}} & {\color{OliveGreen}\textbf{+0.9}} \\
        \bottomrule
    \end{tabular}
    }
\egroup
% \vspace{-0.2cm}
\tiny
\caption{\textbf{Accuracy on Scene Text Benchmarks With and Without using Synthetic Data.} Utilizing the large synthetic datasets of MJ~\cite{jaderberg2014synthetic} and ST~\cite{gupta2016synthetic} improves performance on the more common benchmarks of SVT, IC13, and IC15. However, the averaged performance across all datasets is marginally better due to the existence of many real-world images.}
\label{table:synthetic}
% \vspace{-0.8cm}
\end{center}
\end{table*}

%% file: Tables/uber_category.tex
\begin{table*}[t!]
\normalsize
\begin{center}
\footnotesize
\bgroup
% \def\arraystretch{1.1}
%         \resizebox{1\textwidth}{!}{%
%         \begin{tabular}{lccccccccccc}
%         \toprule
%         \textbf{Category} & \textbf{Business name} & \textbf{Phone number} & \textbf{Street name}& \textbf{None} & \textbf{License plate} & \textbf{Secondary unit designator} & \textbf{Street number} & \textbf{Street number Range} & \textbf{Traffic sign} \\
%         \midrule
%         Size & 14254 & 32 & 5885 & 4866 & 1 & 98 & 22701 & 1708 & 16 \\
%         \midrule
%         Parseq & 85.7 & 50 & 95 & 82.4 & 0 & 86.7 & 78.3 & 96.3 & 93.8 \\ \midrule
%         CLIPTER & 87 & 46.9 & 95.4 & 83.7 & 0 &88.8 & 79.6 & 96.5 & 93.8 \\
%         \bottomrule
%     \end{tabular}
%     }
    \begin{tabular}{lccccccccccc}
        \toprule
        & \textbf{Street} & \textbf{Business} & \textbf{Street}& \multirow{2}{*}{\textbf{None}} & \textbf{Street Number} & \textbf{Secondary Unit} & \textbf{Phone} & \textbf{Traffic} & \textbf{License} \\
        & \textbf{Number} & \textbf{Name} & \textbf{Name} & & \textbf{Range} & \textbf{Designator} & \textbf{Number} & \textbf{Sign} & \textbf{Plate} \vspace{0.1cm} \\
        & 22,701 & 14,254 & 5,885 & 4,866 & 1,708 & 98 & 32 & 16 & 1 \\
        \midrule
        \rowcolor{LightGray} Parseq & 78.3 & 85.7 & 95 & 82.4 & 96.3 & 86.7 & 50 & 93.8 & 0 \\
        + $\text{\AlgoName}_{\text{Vision}}$ & 79.6 & 87 & 95.4 & 83.7 & 96.5 & 88.8 & 46.9 & 93.8 & 0 \\
        $\quad \quad \Delta$ & {\color{OliveGreen}\textbf{+1.3}} & {\color{OliveGreen}\textbf{+1.3}} & {\color{OliveGreen}\textbf{+0.4}} & {\color{OliveGreen}\textbf{+1.3}} & {\color{OliveGreen}\textbf{+0.2}} & {\color{OliveGreen}\textbf{+2.1}} & {\color{BrickRed}-3.1} & 0 & 0 \\
        \bottomrule
    \end{tabular}
\egroup
% \vspace{-0.2cm}
% \tiny
\caption{\textbf{Accuracy on Uber-Text per Word Category.} The number of words in each category is listed below its name. CLIPTER is mostly effective on street numbers and business names, often critical information for scene understanding.}
\label{tab:uber}
% \vspace{-0.8cm}
\end{center}
\end{table*}